\documentclass[sigconf]{aamas}  

\usepackage{amsmath,amsfonts,bm}

\def\eqref#1{equation~\ref{#1}}

\def\1{\bm{1}}

\def\vomega{{\bm{\omega}}}

\def\vr{{\bm{r}}}

\def\vy{{\bm{y}}}
\def\vz{{\bm{z}}}
\def\vG{{\bm{G}}}

\def\vR{{\bm{R}}}
\def\vQ{{\bm{Q}}}

\def\mI{{\bm{I}}}

\DeclareMathAlphabet{\mathsfit}{\encodingdefault}{\sfdefault}{m}{sl}
\SetMathAlphabet{\mathsfit}{bold}{\encodingdefault}{\sfdefault}{bx}{n}

\usepackage{balance} 
\usepackage{multirow}
\usepackage{subfigure}
\usepackage{threeparttable}
\usepackage{algorithm}
\usepackage{algpseudocode}
\usepackage{bbm}

\renewcommand{\boxed}[1]{\bm{#1}}
\newcommand{\miniboxed}[1]{\scalebox{0.6}{#1}}

\setcopyright{ifaamas}
\acmConference[AAMAS '24]{Proc.\@ of the 23rd International Conference
on Autonomous Agents and Multiagent Systems (AAMAS 2024)}{May 6 -- 10, 2024}
{Auckland, New Zealand}{N.~Alechina, V.~Dignum, M.~Dastani, J.S.~Sichman (eds.)}
\copyrightyear{2024}
\acmYear{2024}
\acmDOI{}
\acmPrice{}
\acmISBN{}

\acmSubmissionID{326}

\title[Policy-regularized Offline Multi-objective \\Reinforcement Learning]{Policy-regularized Offline Multi-objective \\ Reinforcement Learning}

\author{Qian Lin}
\affiliation{
  \institution{Sun Yat-sen University}
  \city{Guangzhou}
  \country{China}}
\email{linq67@mail2.sysu.edu.cn}

\author{Chao Yu}
\affiliation{
  \institution{Sun Yat-sen University}
  \city{Guangzhou}
  \country{China}}
\email{yuchao3@mail.sysu.edu.cn}

\author{Zongkai Liu}
\affiliation{
  \institution{Sun Yat-sen University}
  \city{Guangzhou}
  \country{China}}
\email{liuzk@mail2.sysu.edu.cn}

\author{Zifan Wu}
\affiliation{
  \institution{Sun Yat-sen University}
  \city{Guangzhou}
  \country{China}}
\email{wuzf5@mail2.sysu.edu.cn}

\begin{abstract}
In this paper, we aim to utilize only offline trajectory data to train a policy for multi-objective RL. We extend the offline policy-regularized method, a widely-adopted approach for single-objective offline RL problems, into the multi-objective setting in order to achieve the above goal. However, such methods face a new challenge in offline MORL settings, namely the \textit{preference-inconsistent demonstration} problem. We propose two solutions to this problem: 1)  filtering out preference-inconsistent demonstrations via approximating behavior preferences, and 2) adopting regularization techniques with high policy expressiveness. Moreover, we integrate the preference-conditioned scalarized update method into policy-regularized offline RL, in order to simultaneously learn a set of policies using a single policy network, thus reducing the computational cost induced by the training of a large number of individual policies for various preferences. Finally, we introduce \textit{Regularization Weight Adaptation} to dynamically determine  appropriate regularization weights for arbitrary target preferences during deployment. Empirical results on various multi-objective datasets demonstrate the capability of our approach in solving offline MORL problems.~\footnote{The code for this paper can be found at \url{https://github.com/qianlin04/PRMORL}} 
\end{abstract}


\newcommand{\BibTeX}{\rm B\kern-.05em{\sc i\kern-.025em b}\kern-.08em\TeX}
\newif\ifshowappendix
\showappendixtrue  

\begin{document}


\pagestyle{fancy}
\fancyhead{}


\maketitle 


\section{Introduction}
In recent years, Reinforcement Learning (RL)~\citep{sutton2018reinforcement} has achieved tremendous successes in solving various challenging problems including robotics~\citep{kaufmann2023champion,hanna2021grounded}, games~\citep{wang2023voyager,vinyals2019grandmaster} and recommendation systems~\citep{zheng2018drn}. 
In the standard RL setting, the primary goal is to optimize a policy that maximizes a cumulative scalar reward.
However, there often exist multiple, potentially conflicting objectives in many real-world applications, which motivates the research of Multi-Objective Reinforcement Learning (MORL)~\citep{liu2014multiobjective}.
In contrast to the characteristic of single-best solution in single-objective settings, the MORL paradigm accommodates multiple optimal policies catering to various preferences that indicate the trade-off among multiple objectives.
Many MORL approaches have been proposed in the past few years~\citep{liu2014multiobjective,mossalam2016multi,abels2019dynamic}.
However, most existing approaches focus on the online setting, and thus are inappropriate for many safety-critical real-world scenarios due to the potential risk from substantial online exploration.
Offline MORL, which aims to leverage only offline datasets for the training of multi-objective policies, holds promise for solving this challenge but has received limited attention and research.

Current offline MORL algorithms either assume that the target preferences during deployment are static and pre-known~\citep{wu2021offline}, or require access to behavior policies~\citep{thomas2021multi} or their preference information
\citep{zhu2023scaling}, which can hinder the applicability of these algorithms in many offline MORL scenarios.
The primary objective of this paper is to leverage trajectory data obtained from preference-unknown behavior policies to train a policy capable of achieving satisfactory performance for arbitrary target preferences during deployment. 
To achieve this goal, we extend the offline policy-regularized methods~\citep{fujimoto2019off,wu2019behavior,kumar2019stabilizing,wang2022diffusion}, a kind of widely studied and effective offline RL approach, to the multi-objective linear-preference setting to address the offline MORL problem.
First, we introduce the \textit{preference-inconsistent demonstration} problem, a challenge encountered when directly applying offline policy-regularized methods to multi-objective datasets collected by behavior policies with diverse preferences.
Specifically, when a policy makes decisions based on a preference $\vomega'$ deviating from the target preference $\vomega$, trajectories generated by this policy can serve as arbitrarily inferior demonstrations under preference $\vomega$, due to conflicts in final decision-making objectives between $\vomega$ and $\vomega'$.
Consequently, these trajectories can misguide policy learning through the behavior cloning term used by offline policy-regularized methods.
To address this problem, one intuitive approach is to directly filter out preference-inconsistent demonstrations from training datasets, which can be achieve by approximating the preferences of behavior policies.
In addition, given that preference-inconsistent demonstrations can increase the difficulty of implicitly modeling behavior policies in offline policy-regularized methods, we mitigate this issue by adopting regularization techniques with high policy expressiveness.

Another challenge is that training an individual policy for each possible target preference can lead to considerable computational cost.
To solve this problem, we integrate the preference-conditioned scalarized update method \citep{abels2019dynamic} into policy-regularized offline RL by employing the vector value estimate, scalarized value update and preference-conditioned actors and critics.
This approach enables the simultaneous learning of a set of policies with a single policy network.
Moreover, since the optimal regularization weight in offline MORL setting can vary across different preferences, \textit{Regularization Weight Adaptation} is proposed to dynamically determine the appropriate regularization weights for each target preference during deployment.
Specifically, we treat behavior cloning as an additional decision-making objective and the regularization weight as the corresponding preference weight, thus incorporating them into the original multi-objective setting during training.
The appropriate regularization weights then can be dynamically adjusted using a limited number of collected trajectories during deployment.

We conduct several experiments on D4MORL~\citep{zhu2023scaling}, an offline MORL dataset designed for continuous control tasks.
The results demonstrate that, compared to the state-of-the-art offline MORL method PEDA~\citep{zhu2023scaling}, our approach achieves competitive or even better performance without using the preference data of behavior policies.
Additionally, since many current offline datasets are collected by behavior policies with a single and fixed preference, it is meaningful to explore the feasibility of utilizing these datasets for MORL.
Therefore, we introduce the MOSB dataset, a multi-objective variant of D4RL~\citep{fu2020d4rl} collected by single-preference behavior policies, which is different from D4MORL where behavior policies are explicitly trained for MORL.
Our approach consistently exhibits outstanding performance on this dataset, highlighting its superiority in terms of generalizability across various offline MORL scenarios.

\section{Related work}
\textbf{MORL}
\quad
Existing work within the domain of MORL~\cite{liu2014multiobjective,vamplew2011empirical} can be broadly categorized into two classes, namely single-policy and multi-policy methods.
Single-policy methods aim to transform a multi-objective problem into a single-objective problem under one specific preference among multiple objectives, and then solve it to obtain a single optimal policy using standard RL approaches~\citep{agarwal2022multi,abdolmaleki2020distributional}.
Conversely, multi-policy methods learn a set of policies to approximate the Pareto front of optimal policies.
To achieve this goal, one can explicitly maintain a finite set of policies by either applying a single-policy algorithm individually for each candidate preference~\citep{roijers2014linear,mossalam2016multi}, or employing evolutionary algorithms to generate a population of diverse policies~\cite{handa2009solving,xu2020prediction}.
However, these methods suffer from low scalability to high-dimensional preference space due to the limited number of learned policies.
To tackle the above challenge, a feasible approach is to simultaneously learn a set of policies represented by a single network~\citep{abels2019dynamic,yang2019generalized,basaklar2022pd}.
Conditioned Network (CN) method~\citep{abels2019dynamic} utilizes a single preference-conditioned network to represent value functions over the entire preference space and applies the scalarized update method~\citep{mossalam2016multi} to update the vector value function.
Based on this method, several works~\citep{yang2019generalized, basaklar2022pd} are proposed to enhance performance while improving sample efficiency during online training.


\vspace{0.1cm}
\noindent\textbf{Policy-regularized Offline RL}
\quad
The main challenge of offline RL is the distribution shift between the state-action pairs in the dataset and those induced by the learned policy~\citep{levine2020offline}. 
Policy regularization~\citep{fujimoto2019off,wu2019behavior,kumar2019stabilizing,wang2022diffusion} is proposed as a simple yet efficient approach to address this issue. 
This method typically incorporates a behavior cloning mechanism into the policy learning to regularize the learned policy stay close towards the behavior policy.
BCQ~\citep{fujimoto2019off} employs a conditional variational auto-encoder to model behavior policies and learns a perturbation model to generate bounded adjustments to the actions of the behavior policy.
BEAR~\cite{kumar2019stabilizing} regularizes the learned policy with the maximum mean discrepancy, which is estimated using multiple samples from learned policy and pre-modeled behavior policy.
BRAC~\cite{wu2019behavior} proposes a general regularization framework and comprehensively evaluates the existing techniques in recent methods.
TD3+BC~\cite{fujimoto2021minimalist} adds maximum likelihood estimation (MLE) as a minimal change to the TD3 algorithm, which exhibits surprising effectiveness in addressing offline RL problems.
Diffusion Q-learning~\cite{wang2022diffusion} is a state-of-the-art offline policy regularization method that leverages the diffusion model as the learned policy while utilizing the training loss of the diffusion policy as a policy regularizer.

\vspace{0.1cm}
\noindent\textbf{Offline MORL}
\quad
There exist a limited number of works that focus on the offline MORL problem.
PEDI~\citep{wu2021offline} utilize the dual gradient ascent and pessimism principle to find an optimal policy that achieves the highest utility under a fixed preference while keeping the vector value constrained within a target set.
MO-SPIBB~\cite{thomas2021multi} adopts the Seldonian framework~\citep{thomas2019preventing} to achieve safe policy improvement under some predefined preferences using offline datasets.
These methods require prior knowledge of target preferences and cannot be adapted to new preferences during deployment.
To learn a single policy that works for all preferences and solve the continuous control problem in MORL settings, PEDA~\citep{zhu2023scaling} introduces D4MORL, a large-scale benchmark for offline MORL, and extends supervised offline RL methods~\citep{chen2021decision,emmons2021rvs} to MORL by directly incorporating the preference information into the input of decision models.

Compared to the above works, we provide an intuitive yet unexplored solution to offline MORL problems by integrating two well-established methods from the online MORL and offline RL domains, i.e., offline policy-regularized methods and the preference-conditioned scalarized update technique, respectively.
The major advantage of our proposed method is that it can respond to arbitrary target preferences without access to behavior policies or their preference information, which is typically unavailable in real-world scenarios.
The empirical results show that our approach demonstrates superior or competitive performance compared with PEDA on both datasets collected by multi-preference behavior polices and by single-preference behavior polices.





\section{Background}
\textbf{Multi-objective RL}
\quad
The problem of multi-objective RL can be formulated as a multi-objective Markov decision process (MOMDP), which is represented by a tuple $(\mathcal S,\mathcal A,\mathcal P,\vr , \gamma, \Omega, f_{\Omega})$ with state space $\mathcal S$, action space $\mathcal A$, transition distribution $\mathcal P(s'|s,a)$, vector reward function $\vr(s,a)$,  discount factor $\gamma\in [0,1]$, space of preferences $\Omega$, and preference functions $f_{\vomega}(\vr)$.
The vector reward function is $\vr(s,a)=[r(s,a)_1,r(s,a)_2,...,r(s,a)_n]^T$ where $n$ is the number of objectives and $r(s,a)_i$ is the reward of the $i^{\text{th}}$ objective. 
The preference function $f_{\vomega}(\vr)$ maps a reward vector $\vr(s,a)$ to a scalar utility by using the given preference $\vomega\in \Omega$.
The expected vector return $\vG^\pi$ of a policy $\pi$ is denoted as the discount sum of reward vectors, i.e., $\vG^\pi=\mathbb E_\pi[\sum_t \gamma^t\vr(s_t,a_t)]$.

In multi-objective problems, there is typically no optimal policy that maximizes all objectives simultaneously due to the inner conflict among different goals.
Instead, a set of non-dominated policies are needed in multi-objective problems. 
A policy $\pi$ is Pareto optimal if it cannot be dominated by any other policy $\pi'$, i.e., no other policy $\pi'$ can achieve a better return in one objective without incurring a worse return than $\pi$ in other objectives.
In this work, we consider the common framework of an MOMDP with linear preferences, i.e., $f_{\vomega}(\vr(s,a))=\vomega^T\vr(s,a)$ where $\vomega$ is a vector indicating the linear preference.
Therefore, given a preference $\vomega$, we should  train a policy $\pi$ that maximizes the expected scalarized return $\vomega^T \vG^\pi=\mathbb E_\pi[\vomega^T\sum_t \gamma^t\vr(s_t,a_t)]$.
If all possible preferences in $\Omega$ are considered, a set of corresponding Pareto policies can be obtained to form an approximation of the Pareto front.

In the context of offline MORL, we assume no access to online interactions with the environment and offline dataset $\mathcal{D}=\{(s, a, s', \vr)\}$ is the only data available for training.
The dataset $\mathcal{D}$ is generated by the behavior policies $\pi_\beta(\cdot|\vomega)$ based on various preferences $\vomega$ of multiple objectives.
Therefore, we define the behavior preference $\vomega_{\tau}$ as the preference of the policy that generates trajectory $\tau$.
Given that the preferences of behavior policies are either unknown or difficult to be explicitly represented in many real-world applications, in this paper, we address the offline MORL problems under the assumption that the real behavior preferences are inaccessible during the policy training process, and 
the goal is set to learn a policy from $\mathcal{D}$ that performs well for arbitrary target preference during deployment.

\vspace{0.1cm}
\noindent\textbf{Preference-conditioned Scalarized Update}
\quad
To extend the single-objetive methods to multi-objective settings, scalarized deep Q-learning~\citep{mossalam2016multi} proposes to learn a vector-valued Q-value $\vQ=[Q_1, Q_2,...,Q_n]$, where $Q_i$ is the Q value w.r.t. $i^{\text{th}}$ objective.
For a specific target preference $\vomega$, $\vQ$ is updated by minimizing the loss:
\begin{align}
    L_{\vQ}=\mathbb{E}_{(s,a,\vr,s')\sim D}\left[|\vy-\vQ(s,a)|^2\right],
\end{align}
where $\vy=\vr+\gamma \vQ(s', \arg\max_{a'}[\vomega^T\vQ(s',a')])$.
To respond to dynamically determined target preferences, \citet{abels2019dynamic} propose a vector-valued Q-network with outputs conditioned on the given preference, i.e., $\vQ(s,a,\vomega)$. 
This method can be extended to actor-critic framework by introducing the preference-conditioned vector critic $\vQ(s,a,\vomega)$ and actor $\pi(a|s,\vomega)$.
In this framework, for each possible target preference $\vomega$, the critic $\vQ(s,a,\vomega)$ is updated by minimizing the loss:
\begin{align}
    L_{\text{critic}}&=\mathbb{E}_{(s,a,\vr,s')\sim D}\left[|\vy-\vQ(s,a,\vomega)|^2\right],
\end{align}
where $\vy=\vr+\gamma \mathbb E_{a'\sim \pi(\cdot|s,\vomega)}\vQ(s', a', \vomega)$.
As for the actor $\pi(a|s,\vomega)$, the loss can be written as:
\begin{align}
    L_{\text{actor}}&=-\mathbb{E}_{s\sim D}\left[\mathbb{E}_{a\sim \pi(\cdot|s,\vomega)} \left[\vomega^T\vQ(s,a,\vomega)\right]\right].
\end{align}

\vspace{0.1cm}
\noindent\textbf{Offline Policy-regularized Methods}
\quad
In offline RL, the errors in value estimation from out-of-distribution actions often lead to the final poor performance.
To address this issue, policy regularization is applied to encourage the learned policy $\pi$ to stay close to the behavior policy $\pi_{\beta}$, thus preventing out-of-distribution actions and avoiding performance degradation.
In this paper, we focus on a common form of policy regularization, where a behavior cloning term is explicitly incorporated into the original actor loss to maximize the action-state value within the actor-critic framework:
\begin{align}
    L_{\text{actor}}=-\mathbb{E}_{(s,a)\sim D}\left[\mathbb{E}_{a'\sim \pi(\cdot|s)}\left[Q(s,a')\right]-\kappa L_{\text{bc}}(s,a)\right],
\end{align}
where $Q(s,a)$ is state-action value learned by critic, $L_{\text{bc}}$ is a behavior cloning term and $\kappa$ is a regularization weight to control the strength of the behavior cloning.
The widely used behavior cloning terms in previous offline RL studies include:
1) \textbf{MSE regularization}~\citep{fujimoto2021minimalist} extends the TD3~\cite{fujimoto2018addressing} algorithm by adding a behavior cloning term in the form of mean square error (MSE) between the actions of the learned policy and the behavior policy in the dataset, i.e., $L_{\text{bc}}(s,a)=\mathbb{E}_{a'\sim \pi(\cdot|s)}\left[\Vert a-a'\Vert^2\right]$; 
2) \textbf{CVAE regularization} utilizes a policy class based on the conditional variational auto-encoder (CVAE) and employs the loss of evidence lower bound as the behavior cloning term, i.e. $L_{\text{bc}}(s,a)=\mathbb{E}_{z\sim q(\cdot|s,a)}\left[\log p(a|s,z)\right]-D_{\text{KL}}(q(z|s,a)||\mathcal N(\bm{0},\mI))$, where $q$ is an encoder and $p$ is a decoder that is used for decision-making; 
and 3): \textbf{Diffusion regularization}~\citep{wang2022diffusion} adopts the diffusion model as the learned policy and the diffusion reconstruction loss as the behavior cloning term, i.e., $\mathbb{E}_{i\sim \mathcal{U},\epsilon\sim \mathcal N(\bm{0},\mI)}\left[\Vert\epsilon-\epsilon_\theta(\sqrt{\overline{\alpha}_i}a+\sqrt{1-\overline{\alpha}_i}\epsilon, s, i)\Vert^2\right]$, where $i$ is the diffusion timestep, $\overline{\alpha}_i$ are pre-defined parameters and $\epsilon_\theta(\cdot)$ is a diffusion model serves as the learned policy.

\section{Policy-regularized Offline MORL}
In this section, we first introduce the \textit{preference-inconsistent demonstration} (PrefID) problem when applying offline policy-regularized methods to multi-objective datasets collected by behavior policies with diverse preferences, and present two approaches to mitigating this problem: 1) filtering out preference-inconsistent demonstrations via approximating behavior preferences and 2) adopting regularization techniques with high policy expressiveness.
Then, a natural integration of policy-regularized offline RL and the preference-conditioned scalarized update method is proposed to simultaneously learn a set of policies using a single policy network.
Finally, we introduce a regularization weight adaptation method to dynamically determine the appropriate regularization weights for each target preference during deployment.


\subsection{The PrefID Problem}
\label{sec:preference_inconsistent}
When employing the offline policy-regularized method to train an individual policy for a specific target preference $\vomega$, we minimize the following loss:
\begin{align}
    L_{\text{actor}}&=-\mathbb{E}_{(s,a)\sim D}\left[\mathbb{E}_{a'\sim \pi(\cdot|s)}\left[Q(s,a')\right]-\kappa L_{\text{bc}}(s,a)\right],\\
    L_{\text{critic}}&=\mathbb{E}_{(s,a,\vr,s')\sim D}\left[(y-Q(s,a'))^2\right]
    \label{eq:single_objective_ac},
\end{align}
where $y=\vomega^T\vr+\gamma \mathbb E_{a'\sim \pi(\cdot|s)} Q(s', a')$ and $Q(s,a')$ is the value function defined as the expectation of total scalar reward $\vomega^T\vr$. 
Despite its effectiveness in solving traditional single-objective offline problems, this approach encounters a new challenge, i.e., the PrefID problem, when applied to multi-objective datasets collected by behavior policies with diverse preferences.
\begin{figure*}[ht!]
\begin{center}
    \vskip -0.2in
    \centerline{\includegraphics[width=\textwidth]{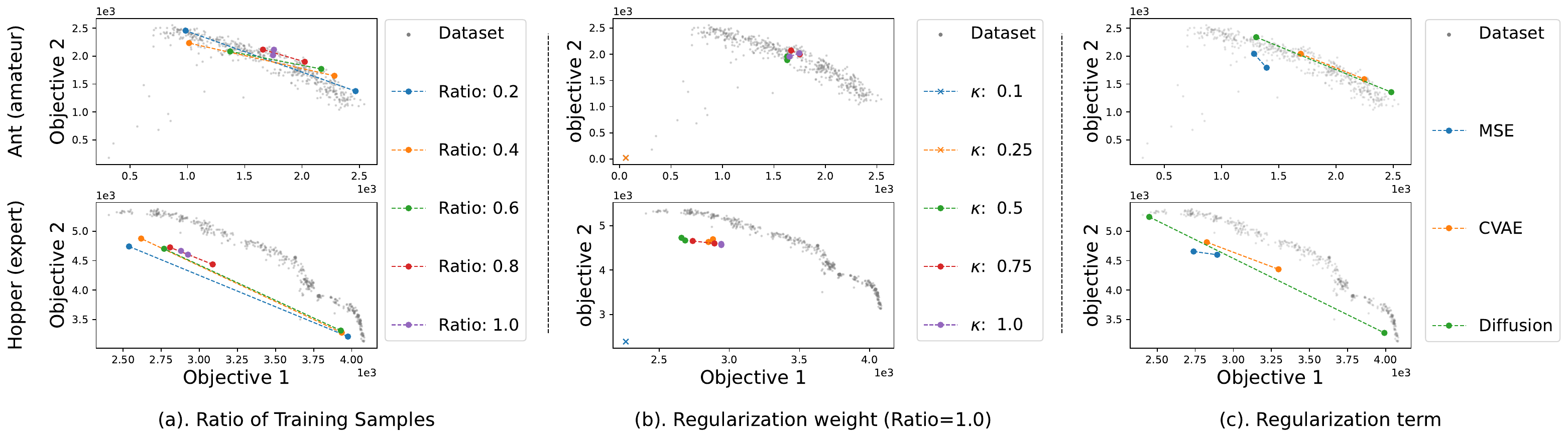}}
    \vskip -0.1in
    \caption{The performance when applying policy-regularized methods on the Ant amateur and Hopper expert datasets within D4MORL.
    Two solid circles with the same color represent the expected vector return of policies trained under preferences $\vomega_1$ and $\vomega_2$ respectively (the farther these two circles extend to the high area of axes, the better performance they exhibit).
    Each cross signifies an aborted training due to the divergence of value estimates. 
    The black dots represent trajectories of entire offline dataset, which reflect performance of behavior policies under various preferences.
    }
    \label{fig:single_obj_exp}
\end{center}
\vskip -0.2in
\end{figure*}
Specifically, if a preference $\vomega'$ deviates from the target preference $\vomega$, the policy based on $\vomega'$ can exhibit a conflict decision-making objective with the policy based on $\vomega$ in decision-making.
Therefore, trajectories generated from behavior preference $\vomega'$ can become arbitrarily inferior demonstrations (i.e., preference-inconsistent demonstrations) during the policy training process under the target preference $\vomega$.
These demonstrations can significantly misguide policy learning through the behavior cloning term used by offline policy-regularized methods and thus lead to performance degradation.



To study the above problem, we first introduce the sub-dataset $\mathcal{D}_{\vomega}$, which is a subset of $\mathcal{D}$ that satisfies: 
\begin{enumerate}
    \item $|\mathcal{D}_{\vomega}|/|\mathcal{D}|=\text{a certain ratio};$
    \item $ 1-\frac{\vomega\cdot\vomega_{\tau_i} }{||\vomega||\cdot||\vomega_{\tau_i} ||}\leq 1-\frac{\vomega\cdot\vomega_{\tau_j} }{||\vomega||\cdot||\vomega_{\tau_j} ||}
    ,\forall \tau_i\in \mathcal{D}_{\vomega},\tau_j\in \mathcal{D}\backslash\mathcal{D}_{\vomega}$.
\end{enumerate}
In other words, $\mathcal{D}_{\vomega}$ contains a specific proportion of offline trajectories with behavior preferences having the smallest cosine distance to preferences $\vomega$ within the entire offline dataset $\mathcal{D}$.
Given that the divergence between two preferences can be effectively captured by their cosine distance under the linear preference setting, $\mathcal{D}_{\vomega}$ with a smaller ratio can be considered to contain fewer preference-inconsistent demonstrations.

To further illustrate the PrefID problem and explore potential solutions, we conduct experiments on D4MORL by utilizing offline policy-regularized methods to train two policies individually, each maximizing the scalar utility under one of the two target preferences $\vomega_1=[1,0]$ and $\vomega_2=[0,1]$.
Figure~\ref{fig:single_obj_exp} (a) shows the performance of the policy trained under $\vomega_i$ using TD3+BC~\cite{fujimoto2021minimalist} on sub-datasets $\mathcal{D}_{\vomega_i}$ with various ratios $|\mathcal{D}_{\vomega_i}|/|\mathcal{D}|$ for $i=1,2$. 
We observe that if the entire dataset is used for policy training (i.e., $|\mathcal{D}_{\vomega_i}|/|\mathcal{D}|$=1.0), the policies learned under preferences $\vomega_1$ and $\vomega_2$ both exhibit inferior performance compared to behavior policies, while obtaining highly similar vector returns.
Moreover, as preference-inconsistent samples are progressively excluded from the training dataset (i.e., $|\mathcal{D}_{\vomega_i}|/|\mathcal{D}|$ decreases), the policy performance gradually improves, which demonstrates that the preference-inconsistent samples in the training dataset can hinder  policy learning.

\subsection{Mitigating the PrefID Problem}
\label{sec:mitigate_prefid}
Based on the above results, one approach to addressing the PrefID problem is to filter out performance-inconsistent demonstrations from the training dataset based on the cosine distance between the behavior preference and the target preference.
However, the real behavior preferences used for calculating cosine distance are inaccessible under our assumption.
Inspired by the observation that the policy preferences and their associated expected vector return often exhibit strong correlation~\cite{basaklar2022pd}, we propose to approximate the behavior preference of trajectories by using the L1 normalization of their vector returns.
Specifically, we assign each trajectory $\tau$ with an approximate behavior preference: 
\begin{align}
    \hat\vomega_\tau=\vR(\tau)/\Vert \vR(\tau)|\Vert_1,
    \label{eq:appro_behavior_pref}
\end{align}
where $\vR(\tau)=\sum_t \vr_t$ is the total vector return of $\tau$.
Then we define the sub-dataset $\mathcal{D}(\vomega)$ used in the policy training  process under the target preference $\vomega$ as:
\begin{align}
    \mathcal{D}(\vomega)=\{\tau|\tau \in \mathcal{D}, 1-\frac{\vomega\cdot\hat{\vomega}_\tau }{||\vomega||\cdot||\hat{\vomega}_\tau ||}\leq 2\theta \},
    \label{eq:Dw}
\end{align}
which indicates that the cosine distance between the approximate behavior preferences in $\mathcal{D}(\vomega)$ and the target preference $\vomega$ does not exceed the preset threshold $2\theta$.
When $\theta$ is set to $1$, i.e., $\mathcal{D}(\vomega)=\mathcal{D}$, the entire offline dataset is employed for policy training under $\vomega$, while only the trajectories with an approximate behavior preference $\hat{\vomega}_\tau=\vomega$ are utilized when $\theta$ is set to $0$.
Although this approach is intuitive and simple, it can effectively mitigate the emergence of preference-inconsistent demonstrations in $\mathcal{D}(\vomega)$. This is because trajectories generated by policies with inconsistent preferences typically exhibit noticeable differences in their vector returns, and thus such trajectories are scarcely included simultaneously in the same $\mathcal{D}(\vomega)$ due to the similarity of $\hat \vomega_\tau$ in $\mathcal{D}(\vomega)$.



In addition to directly excluding performance-inconsistent demonstrations from training datasets, a parallel approach is to mitigate the adverse effect of preference-inconsistent demonstrations from behavior cloning by utilizing a smaller regularization weight.
However, as depicted in Figure~\ref{fig:single_obj_exp} (b), employing MSE regularization with a reduced regularization weight fails to improve policy performance, and even leads to the divergence of value estimates due to distribution shift issues.
Nevertheless, adopting regularization techniques with high policy expressiveness is an effective approach to mitigating the PrefID problem.
Given that preference-inconsistent demonstrations generated from various preferences can exhibit entirely diverse behaviors, the behavior policy that is implicitly modeled by behavior cloning used in offline policy-regularized methods could demonstrate a high level of complexity, which in term leads to inaccurate modeling of the behavior policy and subsequent policy degradation. This issue can be alleviated by employing regularization techniques with high policy expressiveness (e.g., CVAE and diffusion regularization) for policy learning, which have been proven effective in modeling the complex behavior policy~\citep{wang2022diffusion}.
Figure~\ref{fig:single_obj_exp} (c) shows that using diffusion and CVAE regularization can indeed improve performance and the diffusion regularization method performs best among all these regularization techniques.

\begin{algorithm}[t!]
\caption{Policy-regularized Offline MORL}
\label{alg:policy_training}
\label{alg:training}
\begin{algorithmic}[1]
\State \textbf{Input:} Offline dataset $\mathcal{D}$, hyperparameter $\theta$ and $\omega_{\text{bc}}^{\min}$, batch size $B$
\State Initialized policy $\pi$ and critic $\hat\vQ$
\For{each interaction}
    \State Sample mini-batch $\mathcal{B}=\{(s_i,a_i,\vr_i,s'_i)\}_{i=1}^{B}\sim \mathcal{D}$
    \State Sample target preferences $\vomega_i$ from $P(\vomega|\tau_i)$ where $\tau_i$ is the trajectory containing $(s_i,a_i,\vr_i,s'_i)$
    \State Sample $\omega_{\text{bc},i}$ uniformly from $[\omega_{\text{bc}}^{\min}, 1]$ 
    \State Set $\hat \vomega_i=\left((1-\omega_{\text{bc},i})\cdot\vomega, \omega_{\text{bc},i}\right)$
    \State Update the policy by batch gradient descent on $\mathcal{B}$ $$L_{\text{actor}}=-\frac{1}{B}\sum_{i=1}^B\mathbb{E}_{a'\sim \pi(\cdot|s_i,\hat \vomega_i)}\left[\hat \vomega_i\hat \vQ(s_i,a',\hat \vomega_i)\right]$$
    \State Update the critc by batch gradient descent on $\mathcal{B}$ $$L_{\text{critic}}=\frac{1}{B}\sum_{i=1}^B(\vr_i+\gamma \mathbb E_{a'\sim \pi(\cdot|s,\hat \vomega_i)}\vQ(s'_i, a', \hat \vomega_i)-\vQ(s_i,a_i,\hat \vomega_i))^2$$
\EndFor
\end{algorithmic}
\end{algorithm}

\subsection{Responding to Arbitrary Target Preferences}
To respond to arbitrary target preference during deployment, we can utilize offline policy-regularized methods to train an individual policy for each possible preference.
However, this approach faces a significant challenge in terms of computational costs and storage requirements, as it requires the training of a large number of policies to cover all possible preferences.
To address this challenge, we incorporate the preference-conditioned scalarized update method into policy-regularized offline RL, which enables simultaneous learning of a set of policies with a single policy network.
The goal can be formulated as minimizing the loss:
\begin{equation}
    \begin{aligned}
    L_{\text{actor}}=-\mathbb{E}_{
    \vomega\sim P(\Omega)}&\big[\mathbb{E}_{(s,a)\sim D(\vomega)}\\
    &\big[\mathbb{E}_{a'\sim \pi(\cdot|s,\vomega)}\left[ \vomega^T \vQ(s,a',\vomega)\right]-\kappa L_{\text{bc}}(s,a)\big]\big],\\
    L_{\text{critic}}=\mathbb{E}_{ \vomega\sim P(\Omega)}&\left[\mathbb{E}_{(s,a,\vr,s')\sim D(\vomega)}\left[(\vy-\vQ(s,a,\vomega))^2\right]\right]
    \label{eq:multi_objective_ac},
\end{aligned}
\end{equation}
where $\vy=\vr+\gamma \mathbb E_{a'\sim \pi(\cdot|s,\vomega)}\vQ(s', a', \vomega)$, and $P(\Omega)$ is a prior distribution of $\vomega$.
One challenge in the implementation of Eq.~(\ref{eq:multi_objective_ac}) is computational complexity from reconstructing $\mathcal{D}(\vomega)$ for each new $\vomega$ sampled from $ P(\Omega)$ during training  .
Therefore, in our implementation, we adopt an sampling scheme that alters the sampling order of $\tau$ and $\vomega$, i.e., replace $\mathbb{E}_{\vomega\sim P(\vomega)}\mathbb{E}_{\tau\sim D(\vomega)}$ with $\mathbb{E}_{\tau\sim D}\mathbb{E}_{\vomega \sim P(\vomega|\tau)}$ where $P(\vomega|\tau)=U(\{\vomega\in\Omega|1-\frac{\vomega\cdot\hat{\vomega}_\tau }{||\vomega||\cdot||\hat{\vomega}_\tau ||}\leq 2\theta \})$.



In Eq.~(\ref{eq:multi_objective_ac}), $\kappa$ is the regularization weight that determines the trade-off between behavior cloning and policy improvement.
An oversized $\kappa$ can degenerate policy learning into mere behavior cloning, consequently impeding the policy improvement, while an undersized $\kappa$ can lead to the performance degradation due to severe distribution shift issues.
In offline single-objective settings, an appropriate regularization weight can be determined through meticulous fine-tuning. 
However, the optimal regularization weight in offline MORL settings can vary across different preferences, due to the differing data coverage in the optimal trajectory space for various preferences.
Therefore, the regularization weight should be dynamically adjusted to accommodate diverse preferences.

\begin{algorithm}[ht!]
\caption{Regularization Weight Adaption}
\label{alg:policy_adapting}
\label{alg:adaption}
\begin{algorithmic}[1]
\State \textbf{Input:} Target preference $\vomega$, number of update interaction $N$, number of trajectories collected  within each iteration $K$
\State Initialized a truncated Gaussian distribution $\mathcal{N}(\mu,\sigma)$
\For{$i=1,2,..,N$}
    \State Sample $\omega_{\text{bc}}$ from $\mathcal{N}(\mu,\sigma)$ and let $\hat \vomega=\left((1-\omega_{\text{bc}})\cdot\vomega, \omega_{\text{bc}}\right)$
    \State Collect $K$ trajectories via $\pi(\cdot|\hat \vomega)$ to form a mini-batch $\mathcal{B}$
    \State Update $\mu$ and $\sigma$ by minimizing: 
    $$-\mathbb{E}_{\omega_{\text{bc}}\sim \mathcal{N}(\mu,\sigma)}\left[\mathbb{E}_{\tau\sim \mathcal{B}}\left[\vomega\vR(\tau)\right]\right].$$
\EndFor
\State Return $\mu$ as final regularization weight $\omega_{\text{bc}}$
\end{algorithmic}
\end{algorithm}

To this end, we propose \textit{Regularization Weight Adaptation} to achieve adaptive adjustments of the regularization weight for diverse preferences.
First, we define a behavior cloning objective with a scalar value:
\begin{align}
    V_{\text{bc}}^{\vomega}(s)=-\mathbb{E}_{(s',a')\sim D(\vomega)|s'=s }\left[\eta L_{\text{bc}}(s',a')\right],
\end{align}
where $\eta$ is a scale hyperparameter. Then, we introduce its corresponding preference $\omega_{\text{bc}}$ that serves as an adaptable regularization weight.
Next, we incorporate this behavior cloning objective and its corresponding preference into the original MORL framework by extending the value vector and preference vector to include the behavior cloning objective:
\begin{align}
    \hat \vomega&\triangleq \left((1-\omega_{\text{bc}})\cdot\vomega, \omega_{\text{bc}}\right),\\
    \hat \vQ(s,a,\hat\vomega)&\triangleq \left(\vQ(s,a,\hat\vomega), V_{\text{bc}}^{\hat\vomega}(s)\right).
\end{align}
Due to the incorporation of the behavior cloning objective, both of the actor and the critic should be conditioned on an argumented preference $\hat \vomega$, i.e., $\pi(\cdot|s,\hat \vomega)$, $\vQ(s,a,\hat \vomega)$.
Afterwards, the loss of actor and critic can be reformulated as follows:
\begin{equation}
    \begin{aligned}
    L_{\text{actor}}&=-\mathbb{E}_{\hat\vomega\sim  P(\hat\Omega)}\left[\mathbb{E}_{s\sim \mathcal{D}(\vomega)}\left[\mathbb{E}_{a\sim \pi(\cdot|s,\hat \vomega)}\left[\hat \vomega^T\hat \vQ(s,a,\hat \vomega)\right]\right]\right],\\
    L_{\text{critic}}&=\mathbb{E}_{\hat\vomega\sim  P(\hat\Omega)}\left[\mathbb{E}_{(s,a,\vr,s')\sim D(\vomega)}\left[(\vy-\vQ(s,a,\hat \vomega))^2\right]\right].
    \end{aligned}
    \label{eq:multi_objective_ac_dybc}
\end{equation}
where $\vy=\vr+\gamma \mathbb E_{a'\sim \pi(\cdot|s,\hat \vomega)}\vQ(s', a', \hat \vomega)$.
The argumented preference $\hat \vomega$ is obtained by combining $\vomega$ sampled from $P(\vomega|\tau)$ and $\omega_{\text{bc}}$ sampled from $[\omega_{\text{bc}}^{\min}, 1]$ where $\omega_{\text{bc}}^{\min}>0$ is a hyperparameter to circumvent instances of a zero value for $\omega_{\text{bc}}$.
The main discrepancy between Eq.~(\ref{eq:multi_objective_ac}) and Eq.~(\ref{eq:multi_objective_ac_dybc}) lies in the regularization weight, which is a preset hyperparameter $\kappa$ in Eq.~(\ref{eq:multi_objective_ac}), but assumed unknown preference $\hat\vomega$ in Eq.~(\ref{eq:multi_objective_ac_dybc}).

The remaining challenge lies in determining the regularization weight (i.e., behavior cloning preference) $\omega_{\text{bc}}$ for a given target preference $\vomega$ to maximize the expected scalarized return $\vomega \vG^\pi$.
Drawing inspiration from the policy adaptation method in \cite{yang2019generalized}, we employ a gradient approach to efficiently infer the regularization weight $\omega_{\text{bc}}$.
To respond to a given preference $\vomega$ during deployment, we sample $\omega_{\text{bc}}$ from a truncated Gaussian distribution $\mathcal{N}(\mu,\sigma)$ with learnable parameter $\mu$ and $\sigma$, and then combine $\omega_{\text{bc}}$ and $\vomega$ as inputs to the policy $\pi(\cdot|\hat \vomega)$ for decision-making.
During policy deployment, we can iteratively optimize the parameter $\mu$ and $\sigma$ through gradient decent using a limited number of trajectory samples collected by $\pi(\cdot|\hat \vomega)$ as follows:
\begin{align}
    \mathop{\arg\max}_{\mu, \sigma}\mathbb{E}_{\omega_{\text{bc}}\sim \mathcal{N}(\mu,\sigma)}\left[\mathbb{E}_{\tau\sim \pi(\cdot|\hat \vomega)}\left[\vomega^T\vR(\tau)\right]\right].
\end{align}
Despite the involvement of online interactions during the inference of $\omega_{\text{bc}}$, the associated risks are constrained by two key factors: 1) the policy $\pi(\cdot|\hat \vomega)$ is well-trained to encompass a broad range of $\omega_{\text{bc}}$; and 2) only a limited number of samples (e.g. $30$ in our experiments) are sufficient to optimize the two scalar parameters $\mu$ and $\sigma$ to achieve satisfactory performance.
The pseudo codes of policy training and regularization weight adaptation are presented in Algorithm \ref{alg:policy_training} and \ref{alg:policy_adapting}, respectively. 

\section{Experiments}

In this section, we evaluate our approach on D4MORL~\citep{zhu2023scaling}, an offline MORL dataset collected by  behavior policies with diverse preference, and a multi-objective variant of D4RL~\citep{fu2020d4rl} generated by single-objective behavior policies (i.e., the MOSB dataset).
Due to space limits, we present ablation experiments on the technique of filling out the preference-inconsistent samples and \textit{Regularization Weight Adaptation} in Appendix E.1 and E.2, respectively.

\textbf{Evaluation Protocols}
\quad
Given the fact that the true Pareto front is often unavailable in many problems, Hypervolume (\textbf{Hv}) and Sparsity (\textbf{Sp}) are commonly used metrics for evaluating the relative quality of the approximated Pareto front of different algorithms, without requiring knowledge of the true Pareto front~\citep{xu2020prediction,basaklar2022pd,zhu2023scaling,abdolmaleki2020distributional,reymond2022pareto}.
Hv measures the volume enclosed by the returns of policies in the approximated Pareto set $P$, indicating the performance increase across all objectives, while Sp measures the average return distance of two adjacent solutions, offering insights into the density of polices in the approximated Pareto set.
A larger Hv and a smaller Sp indicate a better performance, with Hv typically serving as the primary comparison metric and Sp the secondary metric.
This is because optimizing policies primarily based on Sp metric can lead to very narrow Pareto front and a low Hv.
More details about metrics of Hv and Sp can be found in Appendix C. 
In all experiments, we evaluate the methods using $101$ equidistant preference point pairs (i.e., $[0.00,1.00], [0.01,0.99],...,[1.00, 0.00]$) and average the vector returns of $5$ trajectories sampled from randomly initialized environments for each preference point.
Hv and Sp metrics are calculated based on these averaged vector return.
We perform $3$ runs with various seeds for each algorithm and report the average.
Except for Hv and Sp metrics, Appendix E.3 provides additional performance comparisons using the expected utility (EU) metric. 

\vspace{0.1cm}
\noindent\textbf{Comparative Algorithms}
\quad
We present the results of our proposed policy-regularized offline MORL method with MSE regularization, CVAE regularization and Diffusion regularization.
Our primary comparative baseline is the state-of-the-art method 
\textbf{PEDA}~\citep{zhu2023scaling},
which extends the DT~\citep{chen2021decision} and RvS~\citep{emmons2021rvs} algorithms to MORL settings, referred as \textbf{MODT (P)} and \textbf{MORvS (P)}, respectively, by incorporating preference information and vector-valued returns into corresponding model architectures.
In addition, we also provide the results of the following two methods: 1) preference-conditioned behavior cloning, denoted as \textbf{BC (P)}, where the policy receives not only the state but also its corresponding behavior preference as input and generates predicted actions; and 2) \textbf{MO-CQL}, where the Conservative Q-Learning (CQL)~\citep{kumar2020conservative} is incorporated into the naive scalarized-updated actor-critic to cope with the distribution shift problem.
Note that algorithms MODT (P), MORvS (P) and BC (P) require access to the preference information of behavior policies, whereas our approach and MO-CQL have no such an assumption.
More implementation details of algorithms can be found in Appendix A and B.

\vspace{0.1cm}
\noindent\textbf{Regularization Weight Adaptation Phase}
\quad
Before evaluation, we conduct regularization weight adaptation to infer the appropriate regularization weight for each evaluated preference in our approach. 
In this phase, we employ gradient descent to update the parameters $\mu$ and $\sigma$ of regularization weight distribution $\mathcal{N}(\mu, \sigma)$. 
Specifically, we conduct a total of 3 gradient descent iterations, with each iteration using 10 trajectories collected by $\pi(\cdot|\hat \vomega)$ from the environment.
Finally, we utilize the updated $\mu$ as the final $\omega_{\text{bc}}$ for decision-making of policy $\pi(\cdot|\hat \vomega)$ during evaluation.

\begin{table*}[t!]
\renewcommand\arraystretch{1.0}
\caption{Results on D4MORL Amateur and Expert datasets. B indicates the performance of behavior policies from PEDA. }
\label{tab:D4MORL_comp}
\begin{center}
\begin{threeparttable} 
\begin{tabular}{llcc|cc|cc|ccc}
~ &\bf Env & metric & \bf B & \bf BC (P) & \bf MO-CQL & \bf MODT (P)~\tnote{$\dagger$} & \bf MORvS (P)~\tnote{$\dagger$} & \bf Diffusion & \bf CVAE & \bf MSE \\
\hline
\multirow{10}*{\rotatebox{90}{Amateur}} & \multirow{2}*{Ant} & HV ($10^6$) & $5.61$ & $3.39$\miniboxed{$\pm .81$} & \boxed{$6.28$}\miniboxed{$\pm .00$} & $5.63$\miniboxed{$\pm .03$} & $5.81$\miniboxed{$\pm .03$} & \boxed{$6.15$}\miniboxed{$\pm .01$} & $5.48$\miniboxed{$\pm .01$} & $5.49$\miniboxed{$\pm .02$} \\ 
 ~ & ~ & SP ($10^4$) & - & $2.44$\miniboxed{$\pm 2.27$} & $1.37$\miniboxed{$\pm .02$} & $1.93$\miniboxed{$\pm .27$} & $0.76$\miniboxed{$\pm .08$} & $0.95$\miniboxed{$\pm .03$} & $1.35$\miniboxed{$\pm .38$} & $1.81$\miniboxed{$\pm .19$} \\ 
 \cline{2-11} 
~ & \multirow{2}*{Swimmer} & HV ($10^4$) & $2.11$ & $2.79$\miniboxed{$\pm .03$} & \boxed{$2.94$}\miniboxed{$\pm .03$} & $0.61$\miniboxed{$\pm .01$} & $2.79$\miniboxed{$\pm .05$} & \boxed{$3.11$}\miniboxed{$\pm .03$} & \boxed{$3.14$}\miniboxed{$\pm .01$} & \boxed{$3.02$}\miniboxed{$\pm .03$} \\ 
 ~ & ~ & SP ($10^1$) & - & $1.44$\miniboxed{$\pm .09$} & $11.75$\miniboxed{$\pm .78$} & $1.80$\miniboxed{$\pm .08$} & $1.60$\miniboxed{$\pm .40$} & $3.13$\miniboxed{$\pm .42$} & $3.86$\miniboxed{$\pm .62$} & $3.20$\miniboxed{$\pm .62$} \\ 
 \cline{2-11} 
~ & \multirow{2}*{HalfCheetah} & HV ($10^6$) & \boxed{$5.68$} & $5.42$\miniboxed{$\pm .05$} & \boxed{$5.59$}\miniboxed{$\pm .01$} & \boxed{$5.63$}\miniboxed{$\pm .05$} & \boxed{$5.74$}\miniboxed{$\pm .00$} & \boxed{$5.72$}\miniboxed{$\pm .02$} & \boxed{$5.74$}\miniboxed{$\pm .00$} & \boxed{$5.75$}\miniboxed{$\pm .01$} \\ 
 ~ & ~ & SP ($10^4$) & - & $1.21$\miniboxed{$\pm .58$} & $4.11$\miniboxed{$\pm .16$} & $0.46$\miniboxed{$\pm .05$} & $0.34$\miniboxed{$\pm .06$} & $0.53$\miniboxed{$\pm .21$} & $0.51$\miniboxed{$\pm .08$} & $0.57$\miniboxed{$\pm .09$} \\ 
 \cline{2-11} 
~ & \multirow{2}*{Hopper} & HV ($10^7$) & \boxed{$1.97$} & $1.60$\miniboxed{$\pm .03$} & \boxed{$1.84$}\miniboxed{$\pm .00$} & $1.56$\miniboxed{$\pm .19$} & $1.68$\miniboxed{$\pm .07$} & \boxed{$1.88$}\miniboxed{$\pm .03$} & $1.75$\miniboxed{$\pm .05$} & \boxed{$1.83$}\miniboxed{$\pm .07$} \\ 
 ~ & ~ & SP ($10^5$) & - & $0.21$\miniboxed{$\pm .03$} & $4.57$\miniboxed{$\pm 3.88$} & $2.13$\miniboxed{$\pm 2.15$} & $0.47$\miniboxed{$\pm .16$} & $0.09$\miniboxed{$\pm .03$} & $0.17$\miniboxed{$\pm .03$} & $0.42$\miniboxed{$\pm .15$} \\ 
 \cline{2-11} 
~ & \multirow{2}*{Walker2d} & HV ($10^6$) & \boxed{$4.99$} & $3.67$\miniboxed{$\pm .09$} & $4.15$\miniboxed{$\pm .03$} & $3.17$\miniboxed{$\pm .12$} & \boxed{$4.93$}\miniboxed{$\pm .05$} & \boxed{$4.97$}\miniboxed{$\pm .02$} & \boxed{$4.94$}\miniboxed{$\pm .02$} & \boxed{$4.99$}\miniboxed{$\pm .07$} \\ 
 ~ & ~ & SP ($10^4$) & - & $18.67$\miniboxed{$\pm 16.26$} & $33.14$\miniboxed{$\pm 15.28$} & $46.88$\miniboxed{$\pm 11.87$} & $0.84$\miniboxed{$\pm .12$} & $1.06$\miniboxed{$\pm .17$} & $0.99$\miniboxed{$\pm .15$} & $0.91$\miniboxed{$\pm .27$} \\ 
 \hline
\hline
\multirow{10}*{\rotatebox{90}{Expert}} & \multirow{2}*{Ant} & HV ($10^6$) & \boxed{$6.32$} & $4.42$\miniboxed{$\pm .57$} & \boxed{$6.12$}\miniboxed{$\pm .01$} & $5.89$\miniboxed{$\pm .03$} & $6.08$\miniboxed{$\pm .04$} & \boxed{$6.16$}\miniboxed{$\pm .02$} & $5.72$\miniboxed{$\pm .05$} & $5.75$\miniboxed{$\pm .03$} \\ 
 ~ & ~ & SP ($10^4$) & - & $4.08$\miniboxed{$\pm 2.78$} & $1.55$\miniboxed{$\pm .21$} & $2.33$\miniboxed{$\pm .65$} & $1.15$\miniboxed{$\pm .12$} & $0.71$\miniboxed{$\pm .08$} & $2.04$\miniboxed{$\pm 1.19$} & $1.51$\miniboxed{$\pm .13$} \\ 
 \cline{2-11} 
~ & \multirow{2}*{Swimmer} & HV ($10^4$) & \boxed{$3.25$} & \boxed{$3.18$}\miniboxed{$\pm .01$} & $2.26$\miniboxed{$\pm .04$} & \boxed{$3.22$}\miniboxed{$\pm .01$} & \boxed{$3.22$}\miniboxed{$\pm .01$} & \boxed{$3.19$}\miniboxed{$\pm .01$} & \boxed{$3.14$}\miniboxed{$\pm .03$} & \boxed{$3.15$}\miniboxed{$\pm .02$} \\ 
 ~ & ~ & SP ($10^1$) & - & $4.00$\miniboxed{$\pm .24$} & $0.02$\miniboxed{$\pm .01$} & $2.96$\miniboxed{$\pm .17$} & $4.38$\miniboxed{$\pm .26$} & $5.44$\miniboxed{$\pm 1.25$} & $11.26$\miniboxed{$\pm 4.24$} & $10.60$\miniboxed{$\pm 4.41$} \\ 
 \cline{2-11} 
~ & \multirow{2}*{HalfCheetah} & HV ($10^6$) & \boxed{$5.79$} & \boxed{$5.60$}\miniboxed{$\pm .03$} & \boxed{$5.65$}\miniboxed{$\pm .02$} & \boxed{$5.65$}\miniboxed{$\pm .00$} & \boxed{$5.75$}\miniboxed{$\pm .00$} & \boxed{$5.76$}\miniboxed{$\pm .00$} & \boxed{$5.75$}\miniboxed{$\pm .00$} & \boxed{$5.76$}\miniboxed{$\pm .00$} \\ 
 ~ & ~ & SP ($10^4$) & - & $0.49$\miniboxed{$\pm .18$} & $1.05$\miniboxed{$\pm .12$} & $0.77$\miniboxed{$\pm .06$} & $0.68$\miniboxed{$\pm .10$} & $0.34$\miniboxed{$\pm .04$} & $0.64$\miniboxed{$\pm .09$} & $0.58$\miniboxed{$\pm .01$} \\ 
 \cline{2-11} 
~ & \multirow{2}*{Hopper} & HV ($10^7$) & \boxed{$2.09$} & $0.88$\miniboxed{$\pm .16$} & $1.71$\miniboxed{$\pm .06$} & $1.79$\miniboxed{$\pm .13$} & \boxed{$1.97$}\miniboxed{$\pm .01$} & $1.67$\miniboxed{$\pm .02$} & $1.76$\miniboxed{$\pm .13$} & \boxed{$1.98$}\miniboxed{$\pm .03$} \\ 
 ~ & ~ & SP ($10^5$) & - & $23.62$\miniboxed{$\pm 11.30$} & $3.47$\miniboxed{$\pm .75$} & $1.21$\miniboxed{$\pm .40$} & $2.05$\miniboxed{$\pm 1.75$} & $6.20$\miniboxed{$\pm 3.85$} & $2.28$\miniboxed{$\pm 1.37$} & $0.83$\miniboxed{$\pm .16$} \\ 
 \cline{2-11} 
~ & \multirow{2}*{Walker2d} & HV ($10^6$) & \boxed{$5.21$} & $2.12$\miniboxed{$\pm .55$} & $2.55$\miniboxed{$\pm .03$} & $5.00$\miniboxed{$\pm .08$} & \boxed{$5.05$}\miniboxed{$\pm .07$} & $4.98$\miniboxed{$\pm .06$} & $4.93$\miniboxed{$\pm .11$} & $4.91$\miniboxed{$\pm .04$} \\ 
 ~ & ~ & SP ($10^4$) & - & $34.77$\miniboxed{$\pm 23.21$} & $0.00$\miniboxed{$\pm .00$} & $2.12$\miniboxed{$\pm .60$} & $1.86$\miniboxed{$\pm .51$} & $4.14$\miniboxed{$\pm 1.39$} & $3.67$\miniboxed{$\pm 1.71$} & $3.71$\miniboxed{$\pm 1.38$} \\ 
 \hline
\end{tabular}

\begin{tablenotes}    
    \footnotesize              
    \item[$\dagger$] Due to slight differences in evaluation protocols, the presented results of PEDA exhibit minor discrepancies compared to the original paper.
\end{tablenotes}     
\end{threeparttable}
\end{center}
\end{table*}

\begin{figure*}[ht!]
\vspace{-0.3cm}
\begin{center}
    \centerline{\includegraphics[width=\textwidth]{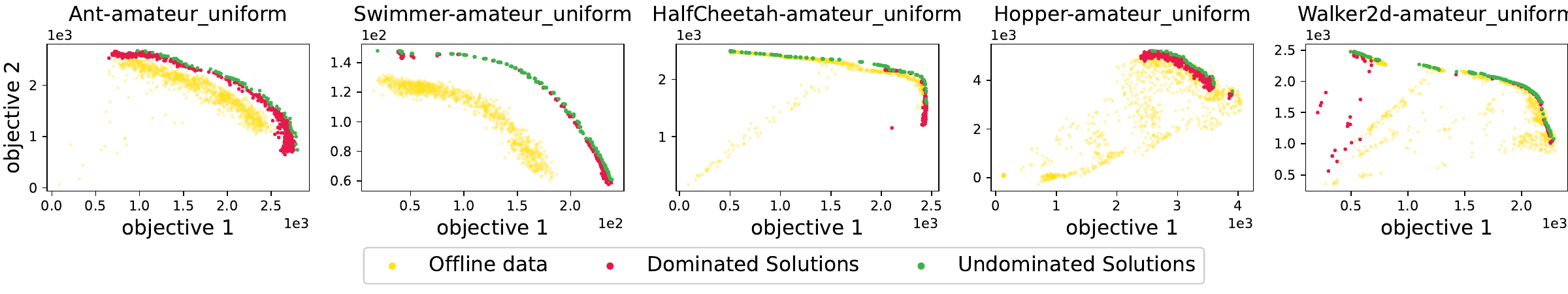}}
    \caption{Approximate Pareto fronts learned by our approach with Diffusion regularization on the Amateur datasets.
    }
    \label{fig:d4morl_pareto_front}
\end{center}
\end{figure*}

\subsection{Experiments on D4MORL}
\label{sec:exp_d4morl}
\textbf{Environments and Datasets}
\quad
D4MORL~\cite{zhu2023scaling} is a dataset designed for offline MORL, which contains millions of trajectories collected by a set of multi-objective behavior policies in various multi-objective MuJoCo environments.
D4MORL contains two types of datasets with different behavior policies: 1) \textbf{Amature Dataset}, a combination of expert demonstrations and suboptimal data generated via randomly perturbed expert policies; and 2) \textbf{Expert Dataset}, consisting entirely of trajectories collected by expert policies.
In addition, the datasets we utilized for comparison consist of trajectories with behavior preferences uniformly sampled from a wide range of preference space.
These datasets also serve as the primary dataset for experimentation in PEDA~\cite{zhu2023scaling}.
More details about D4MORL are provided in Appendix D.


\vspace{0.1cm}
\noindent\textbf{Results and Analysis}
Tables \ref{tab:D4MORL_comp} provides the performance of different methods on the Amateur dataset and Expert dataset.
First, we observe that three regularization techniques yield similar performance, but Diffusion regularization demonstrates an overall advantage compared to the others, which is consistent with the discussions about policy regularization techniques in Section~\ref{sec:mitigate_prefid}.
On amateur dataset, our approach consistently performs the best for the primary metric Hv, especially in the Ant, Simmer and Hopper environments, where our approach achieves significant improvement compared to the state-of-the-art baseline PEDA.
This is because, in contrast to the supervised offline models utilized in PEDA, which strongly rely on high-quality demonstrations for decision-making, our approach utilizes explicit policy improvement through the guidance of value estimates rather than solely following the expert demonstrations, making it more robust to suboptimal demonstrations.
In the Expert dataset, where the data are entirely collected by expert behavior policies, our approach still exhibits competitive performance compared to PEDA.
Finally, we observe that our approach performs slightly worse than PEDA in the secondary metric, Sp. 
A reasonable explanation is that, given the fixed number of sampling preferences, the improvement in Hv, signifying the expansion of the Pareto front, can inevitably lead to increased distances between solutions, consequently leading to an increase in Sp.
The approximate Pareto fronts are partially shown in Figure~\ref{fig:d4morl_pareto_front} and fully displayed in Appendix E.5, which demonstrate the capability of our approach in learning a broad and dense approximate Pareto front.

\subsection{Experiments on MOSB}
\label{sec:exp_d4rl}
\textbf{Environments and Datasets}
\quad
In many real-world scenarios, the offline data are collected by single-objective policies or policies with a fixed preference over multiple objectives. 
To explore the feasibility of utilizing these datasets for MORL, we introduce a new offline multi-objective dataset, namely MOSB, which is built upon the D4RL~\citep{fu2020d4rl} Mujoco dataset.
The main differences between the MOSB dataset and the D4MORL dataset include: 1) behavior preferences, where the behavior preferences in the MOSB dataset are identical and inaccessible, whereas those in D4MORL are sampled from a wide range of preference spaces; 2) behavior policy types, where MOSB is collected from diverse sources, including the partially-trained policy (medium), the replay buffer (medium-replay), and a combination of the partially-trained policy and expert policy (medium-expert); and 3) data quantity, where MOSB contains fewer transition data compared to D4MORL ($0.2\sim 2$M vs. $25$M).
More details about the MOSB dataset are provided in Appendix D.

\begin{table*}[t!]
\caption{Results on MOSB dataset. D indicate the Hv and Sp of the vector return front of offline data.}
\renewcommand\arraystretch{1.0}
\label{tab:mosb_dataset}
\begin{center}
\begin{threeparttable} 
\begin{tabular}{llcc|cc|c|ccc}
~ & \bf Env & \bf Metric & \bf D & \bf BC(P)\tnote{$\dagger$} & \bf MORvS(p)\tnote{$\dagger$} & \bf MO-CQL & \bf Diffusion & \bf CVAE & \bf MSE \\
\hline
\multirow{6}*{\rotatebox{90}{medium-replay}} & \multirow{2}*{HalfCheetah} & Hv ($10^7$) & $1.21$ & $1.56$\miniboxed{$\pm .07$} & $1.71$\miniboxed{$\pm .02$} & $1.59$\miniboxed{$\pm .07$} & \boxed{$3.25$}\miniboxed{$\pm .04$} & \boxed{$3.08$}\miniboxed{$\pm .10$} & $2.93$\miniboxed{$\pm .07$} \\ 
 ~ & ~ & Sp ($10^5$) & $2.74$ & $4.52$\miniboxed{$\pm .46$} & $1.18$\miniboxed{$\pm .36$} & $9.23$\miniboxed{$\pm 1.69$} & $3.32$\miniboxed{$\pm .75$} & $3.55$\miniboxed{$\pm 1.69$} & $3.75$\miniboxed{$\pm 1.17$} \\ 
 \cline{2-10} 
~ & \multirow{2}*{Hopper} & Hv ($10^7$) & $1.22$ & $0.64$\miniboxed{$\pm .15$} & $0.53$\miniboxed{$\pm .03$} & \boxed{$1.37$}\miniboxed{$\pm .07$} & \boxed{$1.46$}\miniboxed{$\pm .01$} & $1.06$\miniboxed{$\pm .08$} & \boxed{$1.27$}\miniboxed{$\pm .06$} \\ 
 ~ & ~ & Sp ($10^5$) & $14.87$ & $7.36$\miniboxed{$\pm 4.89$} & $13.01$\miniboxed{$\pm 4.98$} & $0.59$\miniboxed{$\pm .55$} & $0.48$\miniboxed{$\pm .36$} & $2.76$\miniboxed{$\pm 1.89$} & $0.93$\miniboxed{$\pm .38$} \\ 
 \cline{2-10} 
~ & \multirow{2}*{Walker2d} & Hv ($10^7$) & $1.02$ & $0.42$\miniboxed{$\pm .27$} & $0.96$\miniboxed{$\pm .03$} & $0.37$\miniboxed{$\pm .06$} & \boxed{$1.46$}\miniboxed{$\pm .06$} & $0.95$\miniboxed{$\pm .04$} & $1.17$\miniboxed{$\pm .10$} \\ 
 ~ & ~ & Sp ($10^4$) & $9.71$ & $32.37$\miniboxed{$\pm 3.84$} & $234.56$\miniboxed{$\pm 74.90$} & $190.24$\miniboxed{$\pm 1.04$} & $2.24$\miniboxed{$\pm 1.26$} & $0.08$\miniboxed{$\pm .00$} & $1.04$\miniboxed{$\pm .63$} \\ 
 \hline
\multirow{6}*{\rotatebox{90}{medium}} & \multirow{2}*{HalfCheetah} & Hv ($10^7$) & $1.32$ & $1.17$\miniboxed{$\pm .07$} & $1.17$\miniboxed{$\pm .08$} & $1.19$\miniboxed{$\pm .08$} & \boxed{$3.56$}\miniboxed{$\pm .11$} & $3.18$\miniboxed{$\pm .37$} & \boxed{$3.39$}\miniboxed{$\pm .03$} \\ 
 ~ & ~ & Sp ($10^5$) & $2.29$ & $0.29$\miniboxed{$\pm .22$} & $0.08$\miniboxed{$\pm .05$} & $58.81$\miniboxed{$\pm 1.80$} & $2.08$\miniboxed{$\pm .77$} & $2.19$\miniboxed{$\pm .79$} & $1.00$\miniboxed{$\pm .11$} \\ 
 \cline{2-10} 
~ & \multirow{2}*{Hopper} & Hv ($10^7$) & $1.19$ & $0.43$\miniboxed{$\pm .04$} & $0.52$\miniboxed{$\pm .04$} & \boxed{$1.28$}\miniboxed{$\pm .02$} & \boxed{$1.37$}\miniboxed{$\pm .02$} & \boxed{$1.42$}\miniboxed{$\pm .03$} & \boxed{$1.25$}\miniboxed{$\pm .03$} \\ 
 ~ & ~ & Sp ($10^3$) & $0.00$ & $6.87$\miniboxed{$\pm 2.72$} & $14.33$\miniboxed{$\pm 2.27$} & $29.22$\miniboxed{$\pm 1.96$} & $0.20$\miniboxed{$\pm .05$} & $2.97$\miniboxed{$\pm 1.33$} & $0.78$\miniboxed{$\pm 1.10$} \\ 
 \cline{2-10} 
~ & \multirow{2}*{Walker2d} & Hv ($10^7$) & $0.95$ & $0.64$\miniboxed{$\pm .41$} & $0.97$\miniboxed{$\pm .02$} & $0.80$\miniboxed{$\pm .00$} & $1.27$\miniboxed{$\pm .03$} & \boxed{$1.66$}\miniboxed{$\pm .05$} & $1.17$\miniboxed{$\pm .02$} \\ 
 ~ & ~ & Sp ($10^4$) & $18.82$ & $3.87$\miniboxed{$\pm 2.29$} & $13.65$\miniboxed{$\pm 2.68$} & $1.06$\miniboxed{$\pm .95$} & $0.48$\miniboxed{$\pm .07$} & $0.49$\miniboxed{$\pm .03$} & $0.29$\miniboxed{$\pm .07$} \\ 
 \hline
\multirow{6}*{\rotatebox{90}{medium-expert}} & \multirow{2}*{HalfCheetah} & Hv ($10^7$) & $2.70$ & $1.09$\miniboxed{$\pm .13$} & $2.83$\miniboxed{$\pm .22$} & $2.88$\miniboxed{$\pm .16$} & $3.39$\miniboxed{$\pm .15$} & $2.00$\miniboxed{$\pm .66$} & \boxed{$3.83$}\miniboxed{$\pm .29$} \\ 
 ~ & ~ & Sp ($10^6$) & $1.73$ & $0.26$\miniboxed{$\pm .22$} & $1.73$\miniboxed{$\pm 1.31$} & $0.05$\miniboxed{$\pm .02$} & $2.02$\miniboxed{$\pm .88$} & $1.08$\miniboxed{$\pm .98$} & $0.75$\miniboxed{$\pm .26$} \\ 
 \cline{2-10} 
~ & \multirow{2}*{Hopper} & Hv ($10^7$) & \boxed{$1.48$} & $0.93$\miniboxed{$\pm .45$} & \boxed{$1.54$}\miniboxed{$\pm .01$} & \boxed{$1.54$}\miniboxed{$\pm .01$} & \boxed{$1.56$}\miniboxed{$\pm .01$} & \boxed{$1.52$}\miniboxed{$\pm .01$} & \boxed{$1.48$}\miniboxed{$\pm .01$} \\ 
 ~ & ~ & Sp ($10^2$) & $15.97$ & $9.70$\miniboxed{$\pm 1.37$} & $3.87$\miniboxed{$\pm 1.89$} & $8.04$\miniboxed{$\pm 1.50$} & $2.05$\miniboxed{$\pm .67$} & $1.88$\miniboxed{$\pm 1.15$} & $0.72$\miniboxed{$\pm .44$} \\ 
 \cline{2-10} 
~ & \multirow{2}*{Walker2d} & Hv ($10^7$) & $1.15$ & $1.02$\miniboxed{$\pm .06$} & \boxed{$1.29$}\miniboxed{$\pm .00$} & $1.08$\miniboxed{$\pm .02$} & \boxed{$1.46$}\miniboxed{$\pm .01$} & $0.93$\miniboxed{$\pm .81$} & \boxed{$1.34$}\miniboxed{$\pm .00$} \\ 
 ~ & ~ & Sp ($10^5$) & $9.76$ & $0.44$\miniboxed{$\pm .16$} & $3.74$\miniboxed{$\pm 1.53$} & $1.27$\miniboxed{$\pm 1.26$} & $0.01$\miniboxed{$\pm .00$} & $0.01$\miniboxed{$\pm .01$} & $0.00$\miniboxed{$\pm .00$} \\ 
 \hline
\end{tabular}
\begin{tablenotes}    
    \footnotesize              
    \item[$\dagger$] Since these algorithms require real behavior preferences, which are unavailable in MOSB, we use the approximate preferences in Eq.~(\ref{eq:appro_behavior_pref}) as substitutes for policy training. 
\end{tablenotes}      

\end{threeparttable}
\end{center}
\end{table*}

\begin{figure*}[h!]
\begin{center}
    \centerline{\includegraphics[width=\textwidth]{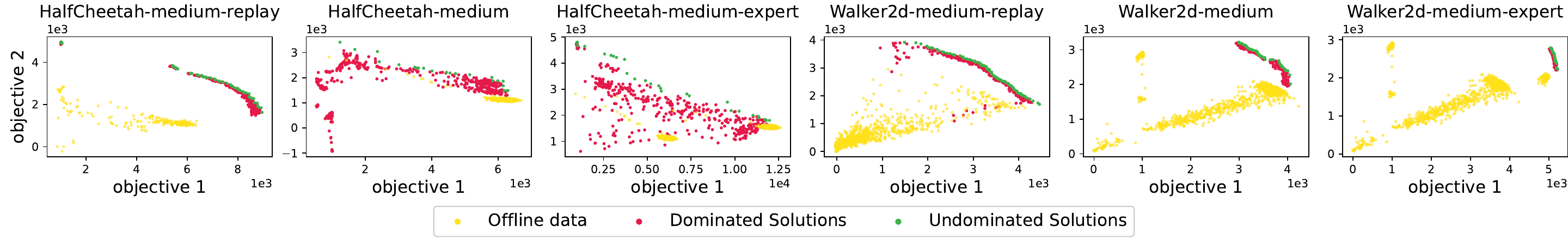}}
    \caption{Vector returns of trajectories in MOSB dataset and approximate Pareto fronts learned by Diffusion regularization.
    }
    \label{fig:d4rl_pareto_front}
\end{center}
\vskip -0.1in
\end{figure*}

\noindent\textbf{Results and Analysis}
\quad
The results of different algorithms on the MOSB dataset are presented in Table~\ref{tab:mosb_dataset}. 
Our approaches equipped with different regularization techniques consistently outperform other methods. 
Compared to the vector return of the offline data (i.e. D in Table~\ref{tab:mosb_dataset}), our approach shows significant improvements in both Hv and Sp metrics.
These improvements indicate that our approach has learned a broadly expanded approximate Pareto front with a dense policy set, which is consistent with the results in Figure~\ref{fig:d4rl_pareto_front}. 
In contrast, BC (P) and PEDA rely on accurate behavior preference data, leading to their poorer performance on MOSB.

\balance

We present vector returns of trajectories in MOSB and the approximate Pareto front learned from our approach on HalfCheetah and Walker2d in Figure~\ref{fig:d4rl_pareto_front}, and a comprehensive display of other environments is provided in Appendix E.5.
Due to the fixed preferences of behavior policies, the vector returns distribution of trajectories in the offline dataset is extremely unbalanced.
There are a greater number of trajectories with high returns on objective 1, while fewer trajectories exhibit high returns on objective 2. 
However, our approach can still achieve a broad approximate Pareto front with a dense policy set, which demonstrates the feasibility of learning multi-objective policies from the dataset collected by single-preference behavior policies.
Moreover, our obtained approximate Pareto front significantly expands outward compared to the front composed of vector returns in the dataset, which indicates that our approach can achieve substantial policy improvement rather than simply imitate behavior policies.

\section{Conclusion}
In this paper, we make a first attempt to employ offline policy-regularized methods to tackle the offline MORL problems.
We provide two intuitive yet effective solutions to mitigating the PrefID problem encountered in such problems.
In addition, we introduce the preference-conditioned scalarized update method to policy-regularized offline RL, in order to simultaneously learn a set of policies using a single policy network.
Moreover, a novel method \textit{Regularization Weight Adaptation} is proposed to dynamically determine the appropriate regularization weights for each target preference during deployment.
Empirically, the experiments conducted on various multi-objective datasets demonstrate that, compared with the state-of-the-art offline MORL algorithms, our approach can achieve competitive or superior performance without requiring preference information of behavior polices.

\balance

\ifshowappendix
\newpage
\appendix
\section{Implementation Details of Policy-regularized Offline MORL}
The pseudo codes for policy training and adaption are presented in Algorithm~\ref{alg:training} and Algorithm~\ref{alg:adaption}. 
We reverse the sampling order for target preferences and offline transition samples, which means that we first collect transition samples and then generate the corresponding target preferences. 
When $\theta=0$, the target preferences are the approximate behavior preferences corresponding to the samples, and when $\theta=1$, the target preferences are randomly sampled from the preference space.
The behavior cloning preferences $\omega_{\text{bc}}$ are sampled uniformly from $[\omega_{\text{bc}}^{\min}, 1]$ to form the augmented preference, where $\omega_{\text{bc}}^{\min}>0$ is a hyperparameter to circumvent instances of a zero value for $\omega_{\text{bc}}$.

Here we list several critical hyperparameters in our experiments on both D4MORL and MOSB. 
These hyperparameters include \textbf{the total training iterations} (set at $1e6$ for both datasets), \textbf{the number of critics} (2 for both datasets), \textbf{the minimum behavior cloning weight} ($\omega_{\text{bc}}^{\min}$ set to 0.2 for both datasets), \textbf{the parameter $\theta$} that determines whether to exclude preference-inconsistent demonstrations (set to 1 for Diffusion regularization on the Amateur dataset and 0 for all other cases in D4MORL, while it's set to 1 in MOSB), \textbf{the behavior cloning scale $\eta$} (in D4MORL, it is set to 100 for Diffusion regularization, 200 for CVAE regularization, and 50 for MSE regularization; in MOSB, it is set to 0.01 for HalfCheetah, 0.5 for Walker2d, and 0.5 for Hopper), \textbf{the update iteration for $\mu$ and $\sigma$} during adaptation (3 iterations for both datasets), and \textbf{the number of trajectories} collected within each iteration during adaptation (set at 10 for both datasets).

\section{Implementation Details of Baselines}
For the baselines MODT (P) and MORvS (P), we use the original hyperparameter setting and implementations in the official codebases, i.e., https://github.com/baitingzbt/PEDA.
For the baseline MO-CQL, we adopt the Gaussian policy and optimize it by minimizing the losses:
\begin{equation}
    \begin{aligned}
    L_{\text{actor}}=&-\mathbb{E}_{
    \vomega\sim P(\Omega)}\big[\mathbb{E}_{(s,a)\sim D(\vomega)}\big[\mathbb{E}_{a'\sim \pi(\cdot|s,\vomega)}\left[ \vomega \vQ(s,a',\vomega)\right]\big]\big],\\
    L_{\text{critic}}=&\ \mathbb{E}_{ \vomega\sim P(\Omega)}\big[\mathbb{E}_{(s,a,\vr,s')\sim D(\vomega)}\big[(\vy-\vQ(s,a,\vomega))^2+\\
    &\alpha\left(\vomega\mathbb{E}_{a'\sim \pi(\cdot|s,\vomega)}\big[\vQ(s,a',\vomega)\big]-\vomega\vQ(s,a,\vomega)\right) \big]\big],
\end{aligned}
\end{equation}
where $\alpha$ is a conservative weight, which is tuned to the value in the set $\{1.0, 2.0, 5.0, 10.0, 20.0\}$ for both D4MORL and MOSB. 

\section{Evaluation Metrics}
\label{sec:evaluation_metrics}
Hypervolume (Hv) and sparsity (Sp) are commonly used metrics for evaluating the relative quality of the approximated Pareto front among different algorithms, when the true Pareto front is unavailable.
Below, we provide the formal definitions of \textit{hypervolume} and \textit{sparsity}, along with respective illustration shown in Figure~\ref{fig:hv_sp}.

\begin{definition}[Hypervolume]
    Hypervolume $\mathcal H(P)$ measures the volume enclosed by the returns of policies in the approximated Pareto set $P$: 
    \begin{align}
        \mathcal H(P)=\int_{R^n}\mathbbm{1}_{H(P,r_0)}(\vz) d\vz,
    \end{align}
    where $H(P,\vr_0)=\{\vz\in R^n|\exists 1\leq i\leq |P|:\vr_0\preceq \vz \preceq P(i)\}$, $r_0$ is the reference point, $P(i)$ is the return of $i^{\text{th}}$ policy in P, $\preceq$ is the dominance relation operator and $\mathbbm{1}_{H(P,r_0)}$ equals $1$ if $\vz\in H(P)$ and $0$ otherwise. 
    Higher hypervolumes are better.
\end{definition}

\begin{definition}[Sparsity]
    Sparsity $\mathcal S(P)$ measures the density of policies in the approximated Pareto set $P$:
    \begin{align}
        \mathcal S(P)=\frac{1}{|P|-1}\sum^{n}_{i=1}\sum^{|P|-1}_{j=1}(\tilde{P}_i(j)-\tilde{P}_i(j+1))^2,
    \end{align}
    where $\tilde{P}_i$ is a list of policies' return sorted by the component w.r.t. $i^{\text{th}}$ objective, $\tilde{P}_i(j)$ is the $j^{\text{th}}$ order of return in sorted list $\tilde{P}_i$.
    Lower sparsity is better.
\end{definition}

\begin{figure}[h]
\begin{center}
\includegraphics[width=0.75\columnwidth]{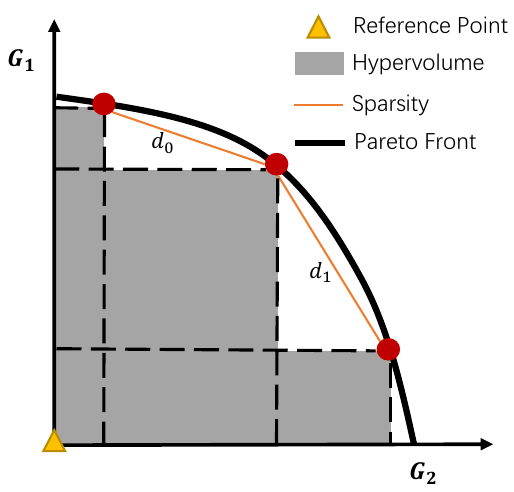}
\end{center}
\caption{Illustration of Hypervolume and sparsity metrics.}
\label{fig:hv_sp}
\end{figure}

\section{Details of Environments and Datasets}


The D4MORL dataset is collected from multi-objective Mujoco environments provided by PGMORL~\cite{xu2020prediction}. 
These environments involve multiple objectives such as running fast, minimizing energy consumption, and achieving high jumping heights. 
One can refer to the PEDA paper~\cite{zhu2023scaling} for more information about the D4MORL dataset.

Our proposed MOSB dataset is built upon the D4RL Mujoco dataset~\cite{fu2020d4rl}, which is collected from single-objective behavior policies. 
we follow the multi-objective setting of D4MORL and set the multiple objectives as running fast and minimizing energy consumption.
We retain the original state and action of transition data in the D4RL Mujoco dataset but recalculate the rewards corresponding to two new objectives for each transition in the dataset thereby constructing vector rewards for subsequent offline MORL.
The distribution of returns on different objectives in the MOSB dataset are depicted by yellow points in Figure~\ref{fig:all_d4rl_pareto_front}. 


\begin{table*}[h]
\caption{D4MORL dataset: the performance of policies trained w or w/o exclusion of preference-inconsistent demonstrations (i.e. $\theta=0$ or $\theta=1$).}
\label{tab:exclude_ablation_d4morl}
\begin{center}
\begin{tabular}{llc|cc|cc|cc}
~ & \bf Env & metric & \bf Diffusion ($\theta$=0) & \bf Diffusion ($\theta$=1) & \bf CVAE ($\theta$=0) & \bf CVAE ($\theta$=1) & \bf MSE ($\theta$=0) & \bf MSE ($\theta$=1) \\
\hline
\multirow{10}*{\rotatebox{90}{Amateur}} & \multirow{2}*{Ant} & Hv ($10^6$) & $6.13$\miniboxed{$\pm .04$} & $6.15$\miniboxed{$\pm .01$} & $5.48$\miniboxed{$\pm .01$} & $4.66$\miniboxed{$\pm .20$} & $5.49$\miniboxed{$\pm .02$} & $3.78$\miniboxed{$\pm .72$} \\ 
 ~ & ~ & Sp ($10^4$) & $0.90$\miniboxed{$\pm .10$} & $0.95$\miniboxed{$\pm .03$} & $1.35$\miniboxed{$\pm .38$} & $0.31$\miniboxed{$\pm .10$} & $1.81$\miniboxed{$\pm .19$} & $0.20$\miniboxed{$\pm .04$} \\ 
 \cline{2-9} 
~ & \multirow{2}*{Swimmer} & Hv ($10^4$) & $3.10$\miniboxed{$\pm .01$} & $3.11$\miniboxed{$\pm .03$} & $3.14$\miniboxed{$\pm .01$} & $2.19$\miniboxed{$\pm .06$} & $3.02$\miniboxed{$\pm .03$} & $0.75$\miniboxed{$\pm .15$} \\ 
 ~ & ~ & Sp ($10^1$) & $12.19$\miniboxed{$\pm 5.35$} & $3.13$\miniboxed{$\pm .42$} & $3.86$\miniboxed{$\pm .62$} & $1.45$\miniboxed{$\pm .47$} & $3.20$\miniboxed{$\pm .62$} & $0.00$\miniboxed{$\pm .00$} \\ 
 \cline{2-9} 
~ & \multirow{2}*{HalfCheetah} & Hv ($10^6$) & $5.69$\miniboxed{$\pm .01$} & $5.72$\miniboxed{$\pm .02$} & $5.74$\miniboxed{$\pm .00$} & $4.54$\miniboxed{$\pm .08$} & $5.75$\miniboxed{$\pm .01$} & $3.54$\miniboxed{$\pm .89$} \\ 
 ~ & ~ & Sp ($10^4$) & $1.41$\miniboxed{$\pm .40$} & $0.53$\miniboxed{$\pm .21$} & $0.51$\miniboxed{$\pm .08$} & $0.01$\miniboxed{$\pm .00$} & $0.57$\miniboxed{$\pm .09$} & $0.00$\miniboxed{$\pm .00$} \\ 
 \cline{2-9} 
~ & \multirow{2}*{Hopper} & Hv ($10^7$) & $1.84$\miniboxed{$\pm .02$} & $1.88$\miniboxed{$\pm .03$} & $1.75$\miniboxed{$\pm .05$} & $1.46$\miniboxed{$\pm .03$} & $1.83$\miniboxed{$\pm .07$} & $1.55$\miniboxed{$\pm .02$} \\ 
 ~ & ~ & Sp ($10^5$) & $1.43$\miniboxed{$\pm 1.02$} & $0.09$\miniboxed{$\pm .03$} & $0.17$\miniboxed{$\pm .03$} & $0.01$\miniboxed{$\pm .01$} & $0.42$\miniboxed{$\pm .15$} & $0.02$\miniboxed{$\pm .00$} \\ 
 \cline{2-9} 
~ & \multirow{2}*{Walker2d} & Hv ($10^6$) & $4.95$\miniboxed{$\pm .01$} & $4.73$\miniboxed{$\pm .01$} & $4.94$\miniboxed{$\pm .02$} & $3.69$\miniboxed{$\pm .21$} & $4.99$\miniboxed{$\pm .07$} & $2.63$\miniboxed{$\pm .13$} \\ 
 ~ & ~ & Sp ($10^4$) & $1.36$\miniboxed{$\pm .07$} & $3.33$\miniboxed{$\pm .79$} & $0.99$\miniboxed{$\pm .15$} & $1.36$\miniboxed{$\pm .42$} & $0.91$\miniboxed{$\pm .27$} & $0.04$\miniboxed{$\pm .03$} \\ 
 \hline
\hline
\multirow{10}*{\rotatebox{90}{Expert}} & \multirow{2}*{Ant} & Hv ($10^6$) & $6.15$\miniboxed{$\pm .02$} & $5.80$\miniboxed{$\pm .09$} & $5.70$\miniboxed{$\pm .04$} & $5.47$\miniboxed{$\pm .07$} & $5.75$\miniboxed{$\pm .04$} & $4.16$\miniboxed{$\pm .55$} \\ 
 ~ & ~ & Sp ($10^4$) & $0.76$\miniboxed{$\pm .04$} & $2.01$\miniboxed{$\pm .58$} & $2.30$\miniboxed{$\pm 1.39$} & $0.59$\miniboxed{$\pm .08$} & $1.50$\miniboxed{$\pm .15$} & $0.51$\miniboxed{$\pm .01$} \\ 
 \cline{2-9} 
~ & \multirow{2}*{Swimmer} & Hv ($10^4$) & $3.19$\miniboxed{$\pm .01$} & $2.83$\miniboxed{$\pm .08$} & $3.13$\miniboxed{$\pm .03$} & $1.02$\miniboxed{$\pm .31$} & $3.16$\miniboxed{$\pm .02$} & $2.63$\miniboxed{$\pm .13$} \\ 
 ~ & ~ & Sp ($10^1$) & $4.58$\miniboxed{$\pm .35$} & $30.08$\miniboxed{$\pm 1.89$} & $11.78$\miniboxed{$\pm 5.12$} & $9.99$\miniboxed{$\pm 9.77$} & $7.79$\miniboxed{$\pm 2.33$} & $1.28$\miniboxed{$\pm .21$} \\ 
 \cline{2-9} 
~ & \multirow{2}*{HalfCheetah} & Hv ($10^6$) & $5.75$\miniboxed{$\pm .00$} & $5.73$\miniboxed{$\pm .00$} & $5.76$\miniboxed{$\pm .00$} & $4.81$\miniboxed{$\pm .11$} & $5.76$\miniboxed{$\pm .00$} & $2.85$\miniboxed{$\pm .81$} \\ 
 ~ & ~ & Sp ($10^4$) & $0.37$\miniboxed{$\pm .00$} & $0.45$\miniboxed{$\pm .09$} & $0.58$\miniboxed{$\pm .05$} & $0.04$\miniboxed{$\pm .04$} & $0.59$\miniboxed{$\pm .01$} & $0.02$\miniboxed{$\pm .00$} \\ 
 \cline{2-9} 
~ & \multirow{2}*{Hopper} & Hv ($10^7$) & $1.46$\miniboxed{$\pm .21$} & $0.96$\miniboxed{$\pm .25$} & $1.84$\miniboxed{$\pm .05$} & $1.63$\miniboxed{$\pm .02$} & $1.98$\miniboxed{$\pm .04$} & $1.63$\miniboxed{$\pm .01$} \\ 
 ~ & ~ & Sp ($10^5$) & $2.07$\miniboxed{$\pm 2.07$} & $4.53$\miniboxed{$\pm .34$} & $3.05$\miniboxed{$\pm 1.01$} & $0.19$\miniboxed{$\pm .11$} & $0.72$\miniboxed{$\pm .07$} & $0.05$\miniboxed{$\pm .02$} \\ 
 \cline{2-9} 
~ & \multirow{2}*{Walker2d} & Hv ($10^6$) & $5.02$\miniboxed{$\pm .02$} & $4.50$\miniboxed{$\pm .01$} & $4.93$\miniboxed{$\pm .14$} & $3.97$\miniboxed{$\pm .19$} & $4.92$\miniboxed{$\pm .04$} & $3.99$\miniboxed{$\pm .03$} \\ 
 ~ & ~ & Sp ($10^4$) & $3.18$\miniboxed{$\pm .34$} & $12.21$\miniboxed{$\pm .57$} & $4.01$\miniboxed{$\pm 2.01$} & $2.52$\miniboxed{$\pm 1.41$} & $2.74$\miniboxed{$\pm .15$} & $0.40$\miniboxed{$\pm .28$} \\ 
 \hline
\end{tabular}
\end{center}
\end{table*}

\begin{table}[h]
\caption{D4MORL dataset: the performance of MO-CQL w or w/o exclusion of preference-inconsistent demonstrations (i.e. $\theta=0$ or $\theta=1$).}
\label{tab:exclude_ablation_mocql}
\begin{center}
\begin{tabular}{llc|cc}
~ & \bf Env & metric & \bf MO-CQL ($\theta$=0) & \bf MO-CQL ($\theta$=1)  \\
\hline
\multirow{10}*{\rotatebox{90}{Amateur}} & \multirow{2}*{Ant} & Hv ($10^6$) & $6.25$\miniboxed{$\pm .04$} & $6.28$\miniboxed{$\pm .00$} \\ 
 ~ & ~ & Sp ($10^4$) & $0.88$\miniboxed{$\pm .11$} & $1.37$\miniboxed{$\pm .02$} \\ 
 \cline{2-5} 
~ & \multirow{2}*{Swimmer} & Hv ($10^4$) & \boxed{$3.16$}\miniboxed{$\pm .01$} & $2.94$\miniboxed{$\pm .03$} \\ 
 ~ & ~ & Sp ($10^1$) & $10.28$\miniboxed{$\pm 3.17$} & $11.75$\miniboxed{$\pm .78$} \\ 
 \cline{2-5} 
~ & \multirow{2}*{HalfCheetah} & Hv ($10^6$) & \boxed{$5.74$}\miniboxed{$\pm .01$} & $5.59$\miniboxed{$\pm .01$} \\ 
 ~ & ~ & Sp ($10^4$) & $0.54$\miniboxed{$\pm .12$} & $4.11$\miniboxed{$\pm .16$} \\ 
 \cline{2-5} 
~ & \multirow{2}*{Hopper} & Hv ($10^7$) & $1.80$\miniboxed{$\pm .01$} & $1.84$\miniboxed{$\pm .00$} \\ 
 ~ & ~ & Sp ($10^5$) & $1.28$\miniboxed{$\pm .06$} & $4.57$\miniboxed{$\pm 3.88$} \\ 
 \cline{2-5} 
~ & \multirow{2}*{Walker2d} & Hv ($10^6$) & \boxed{$4.89$}\miniboxed{$\pm .01$} & $4.15$\miniboxed{$\pm .03$} \\ 
 ~ & ~ & Sp ($10^4$) & $4.62$\miniboxed{$\pm .30$} & $33.14$\miniboxed{$\pm 15.28$} \\ 
 \hline
\multirow{10}*{\rotatebox{90}{Expert}} & \multirow{2}*{Ant} & Hv ($10^6$) & $6.19$\miniboxed{$\pm .02$} & $6.12$\miniboxed{$\pm .01$} \\ 
 ~ & ~ & Sp ($10^4$) & $0.96$\miniboxed{$\pm .11$} & $1.55$\miniboxed{$\pm .21$} \\ 
 \cline{2-5} 
~ & \multirow{2}*{Swimmer} & Hv ($10^4$) & \boxed{$3.21$}\miniboxed{$\pm .01$} & $2.26$\miniboxed{$\pm .04$} \\ 
 ~ & ~ & Sp ($10^1$) & $4.45$\miniboxed{$\pm 1.37$} & $0.02$\miniboxed{$\pm .01$} \\ 
 \cline{2-5} 
~ & \multirow{2}*{HalfCheetah} & Hv ($10^6$) & $5.70$\miniboxed{$\pm .01$} & $5.65$\miniboxed{$\pm .02$} \\ 
 ~ & ~ & Sp ($10^4$) & $0.72$\miniboxed{$\pm .17$} & $1.05$\miniboxed{$\pm .12$} \\ 
 \cline{2-5} 
~ & \multirow{2}*{Hopper} & Hv ($10^7$) & $1.45$\miniboxed{$\pm .00$} & \boxed{$1.71$}\miniboxed{$\pm .06$} \\ 
 ~ & ~ & Sp ($10^5$) & $2.29$\miniboxed{$\pm .00$} & $3.47$\miniboxed{$\pm .75$} \\ 
 \cline{2-5} 
~ & \multirow{2}*{Walker2d} & Hv ($10^6$) & \boxed{$4.54$}\miniboxed{$\pm .05$} & $2.55$\miniboxed{$\pm .03$} \\ 
 ~ & ~ & Sp ($10^4$) & $13.51$\miniboxed{$\pm 1.73$} & $0.00$\miniboxed{$\pm .00$} \\ 
 \hline
\end{tabular}
\end{center}
\end{table}

\section{Additional Results}
\subsection{Ablation on Exclusion of Preference-inconsistent Demonstrations}
Table~\ref{tab:exclude_ablation_d4morl} provide a comprehensive overview of policy performance, with and without the exclusion of preference-inconsistent demonstrations, respectively.
We observed that the excluding preference-inconsistent demonstrations significantly improves the policy performance, especially for MSE regularization and CVAE regularization, which are more vulneratble to the misguidance of preference-inconsistent demonstrations. 
In contrast, the Diffusion regularization method appears to be more resilient to preference-inconsistent demonstrations. 
On the amateur dataset, the Diffusion regularization method with exclusion of preference-inconsistent demonstrations even slightly outperforms the approach without exclusion.

Besides, to further explore the effect of the exclusion technique, we combine MO-CQL with the exclusion of preference-inconsistent demonstrations, and present the comparative performance in Table~\ref{tab:exclude_ablation_mocql}.
The results demonstrate a performance improvement when incorporating the PrefID exclusion into MO-CQL, which verifies the effectiveness of the exclusion technique as well as the potential benefit of combining it with other offline MORL approaches.

\subsection{Ablation on Regularization Weight Adaptation}
To further verify the effectiveness of Regularization Weight Adaptation in improving final performance, we compare the performance of learned policies under different regularization weight settings:
1) \textbf{Fixed}: using a fixed regularization weight; 2) \textbf{Adapting}: evaluating average performance of $30$ samples collected for regularization weight adaptation; 3) \textbf{Adapted}: utilizing the regularization weight obtained via regularization weight adaptation; and 4) \textbf{Oracle}: employing the best-performing regularization weight from $20$ equidistant point within the range of regularization weight.
We use the performance under the \textbf{Oracle} setting to approximate the upper performance bound achievable through tuning the regularization weight. 
Note that the results of our approach presented in the full paper are generated under the \textbf{Adapted} setting. 
The results are presented in Table~\ref{tab:diff_wbc_eval}.

We can observe that policies using the adapted regularization weight exhibit better performance in both Hv and Sp compared to a fixed regularization weight. 
Additionally, the stable performance under the \textbf{Adapting} setting indicates that there is no significant performance degradation during regularization weight adaptation, thereby avoiding additional risks from online interactions.
Finally, the outstanding performance under the \textbf{Oracle} setting demonstrates the potential of our approach for achieving better performance with an increase in the deployment time.


\begin{table}[ht!]
\setlength{\tabcolsep}{4pt}
\renewcommand\arraystretch{1.0}
\caption{Results on D4MORL Amateur dataset under different regularization weight settings.}
\label{tab:diff_wbc_eval}
\begin{center}
\begin{tabular}{lccccc}
\bf Env & metric & \bf Fixed & \bf Adapting & \bf Adapted & \bf Oracle  \\
\hline
\multirow{2}*{Ant} & HV ($10^6$) & $6.11$ & $6.12$\miniboxed{$\pm .01$} & $6.15$\miniboxed{$\pm .01$} & $6.23$\miniboxed{$\pm .02$} \\ 
 ~ & SP ($10^4$) & $1.22$ & $0.59$\miniboxed{$\pm .05$} & $0.95$\miniboxed{$\pm .03$} & $0.74$\miniboxed{$\pm .14$} \\ 
 \cline{3-6} 
\multirow{2}*{Swimmer} & HV ($10^4$) & $3.09$ & $3.10$\miniboxed{$\pm .02$} & $3.11$\miniboxed{$\pm .03$} & $3.13$\miniboxed{$\pm .02$} \\ 
 ~ & SP ($10^1$) & $4.32$ & $2.25$\miniboxed{$\pm .37$} & $3.13$\miniboxed{$\pm .42$} & $2.64$\miniboxed{$\pm .19$} \\ 
 \cline{3-6} 
\multirow{2}*{HalfCheetah} & HV ($10^6$) & $5.72$ & $5.72$\miniboxed{$\pm .02$} & $5.72$\miniboxed{$\pm .02$} & $5.73$\miniboxed{$\pm .02$} \\ 
 ~ & SP ($10^4$) & $0.51$ & $0.29$\miniboxed{$\pm .06$} & $0.53$\miniboxed{$\pm .21$} & $0.57$\miniboxed{$\pm .23$} \\ 
 \cline{3-6} 
\multirow{2}*{Hopper} & HV ($10^7$) & $1.82$ & $1.85$\miniboxed{$\pm .04$} & $1.88$\miniboxed{$\pm .03$} & $1.91$\miniboxed{$\pm .03$} \\ 
 ~ & SP ($10^5$) & $0.09$ & $0.03$\miniboxed{$\pm .02$} & $0.09$\miniboxed{$\pm .03$} & $0.08$\miniboxed{$\pm .03$} \\ 
 \cline{3-6} 
\multirow{2}*{Walker2d} & HV ($10^6$) & $4.88$ & $4.79$\miniboxed{$\pm .03$} & $4.97$\miniboxed{$\pm .02$} & $5.02$\miniboxed{$\pm .01$} \\ 
 ~ & SP ($10^4$) & $2.55$ & $4.39$\miniboxed{$\pm .85$} & $1.06$\miniboxed{$\pm .17$} & $0.60$\miniboxed{$\pm .07$} \\ 
 \hline
\end{tabular}
\end{center}
\vskip -0.1in
\end{table}

\begin{table*}[ht!]
\renewcommand\arraystretch{1.0}
\vskip -0.1in
\caption{Expected Utility (EU) performance on D4MORL Amateur and Expert datasets. }
\vskip -0.15in
\label{tab:D4MORL_comp_eu}
\begin{center}
\begin{tabular}{llc|cc|cc|ccc}
~ &\bf Env & metric & \bf BC (P) & \bf MO-CQL & \bf MODT (P)~\tnote{$\dagger$} & \bf MORvS (P)~\tnote{$\dagger$} & \bf Diffusion & \bf CVAE & \bf MSE \\
\hline
\multirow{5}*{\rotatebox{90}{Amateur}} & \multirow{1}*{Ant} & EU ($10^3$) & $1.46$\miniboxed{$\pm .32$} & \boxed{$2.30$}\miniboxed{$\pm .00$} & \boxed{$2.14$}\miniboxed{$\pm .01$} & \boxed{$2.15$}\miniboxed{$\pm .01$} & \boxed{$2.28$}\miniboxed{$\pm .00$} & $2.03$\miniboxed{$\pm .04$} & $2.05$\miniboxed{$\pm .02$} \\ 
 \cline{2-10} 
~ & \multirow{1}*{Swimmer} & EU ($10^3$) & $0.14$\miniboxed{$\pm .00$} & \boxed{$0.16$}\miniboxed{$\pm .00$} & $0.09$\miniboxed{$\pm .00$} & $0.14$\miniboxed{$\pm .00$} & \boxed{$0.17$}\miniboxed{$\pm .00$} & \boxed{$0.16$}\miniboxed{$\pm .00$} & \boxed{$0.15$}\miniboxed{$\pm .00$} \\ 
 \cline{2-10} 
~ & \multirow{1}*{HalfCheetah} & EU ($10^3$) & $2.00$\miniboxed{$\pm .16$} & \boxed{$2.20$}\miniboxed{$\pm .01$} & \boxed{$2.21$}\miniboxed{$\pm .01$} & \boxed{$2.10$}\miniboxed{$\pm .05$} & \boxed{$2.26$}\miniboxed{$\pm .00$} & \boxed{$2.20$}\miniboxed{$\pm .00$} & \boxed{$2.20$}\miniboxed{$\pm .00$} \\ 
 \cline{2-10} 
~ & \multirow{1}*{Hopper} & EU ($10^3$) & $1.10$\miniboxed{$\pm .29$} & $3.21$\miniboxed{$\pm .09$} & $1.41$\miniboxed{$\pm .34$} & $1.40$\miniboxed{$\pm .11$} & \boxed{$4.06$}\miniboxed{$\pm .00$} & $3.21$\miniboxed{$\pm .28$} & $3.34$\miniboxed{$\pm .24$} \\ 
 \cline{2-10} 
~ & \multirow{1}*{Walker2d} & EU ($10^3$) & $1.02$\miniboxed{$\pm .08$} & $1.29$\miniboxed{$\pm .37$} & $0.94$\miniboxed{$\pm .03$} & $1.71$\miniboxed{$\pm .07$} & \boxed{$2.01$}\miniboxed{$\pm .01$} & \boxed{$1.97$}\miniboxed{$\pm .03$} & \boxed{$1.99$}\miniboxed{$\pm .01$} \\ 
 \hline
\hline
\multirow{5}*{\rotatebox{90}{Expert}} & \multirow{1}*{Ant} & EU ($10^3$) & $1.69$\miniboxed{$\pm .27$} & \boxed{$2.26$}\miniboxed{$\pm .01$} & \boxed{$2.18$}\miniboxed{$\pm .02$} & \boxed{$2.19$}\miniboxed{$\pm .01$} & \boxed{$2.27$}\miniboxed{$\pm .00$} & $2.04$\miniboxed{$\pm .09$} & \boxed{$2.08$}\miniboxed{$\pm .04$} \\ 
 \cline{2-10} 
~ & \multirow{1}*{Swimmer} & EU ($10^3$) & \boxed{$0.16$}\miniboxed{$\pm .00$} & \boxed{$0.15$}\miniboxed{$\pm .00$} & \boxed{$0.16$}\miniboxed{$\pm .00$} & \boxed{$0.16$}\miniboxed{$\pm .00$} & \boxed{$0.15$}\miniboxed{$\pm .01$} & \boxed{$0.15$}\miniboxed{$\pm .01$} & \boxed{$0.15$}\miniboxed{$\pm .00$} \\ 
 \cline{2-10} 
~ & \multirow{1}*{HalfCheetah} & EU ($10^3$) & \boxed{$2.16$}\miniboxed{$\pm .01$} & \boxed{$2.26$}\miniboxed{$\pm .01$} & \boxed{$2.20$}\miniboxed{$\pm .01$} & \boxed{$2.18$}\miniboxed{$\pm .00$} & \boxed{$2.25$}\miniboxed{$\pm .00$} & \boxed{$2.19$}\miniboxed{$\pm .00$} & \boxed{$2.20$}\miniboxed{$\pm .01$} \\ 
 \cline{2-10} 
~ & \multirow{1}*{Hopper} & EU ($10^3$) & $0.65$\miniboxed{$\pm .25$} & $2.42$\miniboxed{$\pm .40$} & $1.54$\miniboxed{$\pm .04$} & $1.64$\miniboxed{$\pm .22$} & \boxed{$2.58$}\miniboxed{$\pm .20$} & $1.77$\miniboxed{$\pm .16$} & \boxed{$2.68$}\miniboxed{$\pm .39$} \\ 
 \cline{2-10} 
~ & \multirow{1}*{Walker2d} & EU ($10^3$) & $0.73$\miniboxed{$\pm .12$} & $0.86$\miniboxed{$\pm .04$} & $1.51$\miniboxed{$\pm .08$} & $1.61$\miniboxed{$\pm .01$} & \boxed{$1.86$}\miniboxed{$\pm .06$} & \boxed{$1.82$}\miniboxed{$\pm .13$} & \boxed{$1.83$}\miniboxed{$\pm .10$} \\ 
 \hline
\end{tabular}

\end{center}
\end{table*}

\subsection{Performance on Expected Utility Metric}
The expected utility (EU) metric is calculated by:
\begin{equation}
    \begin{aligned}
    \text{EU}=\mathbb{E}_{
    \vomega\sim P_{\text{deploy}}(\vomega)}\left[\mathbb{E}_{\pi({\cdot|\vomega})}\left[\vomega^T\sum_t \vr(s_t,a_t)\right]\right]
    \label{eq:eu_metric},
\end{aligned}
\end{equation}
where $P_{\text{deploy}}(\vomega)$ is the preference distribution during deployment.
We assume that $P_{\text{deploy}}(\vomega)$ is a uniform distribution on the whole feasible preference space, and then approximate the expectation $\mathbb{E}_{
    \vomega\sim P_{\text{deploy}}(\vomega)}\left[\cdot\right]$ by averaging over 101 equidistant preference vector.
The EU performance on the D4MORL are presented on Table~\ref{tab:D4MORL_comp_eu}.

Table~\ref{tab:D4MORL_comp_eu} demonstrates competitive or superior performance of our algorithm in the EU metric, aligning with the outcomes assessed by Hv and Sp metrics.
We note a significant advantage of our algorithm in Hopper and Walker2d environments.
This is because, in these environments, most solutions derived by PEDA for out-of-distribution target preferences significantly stray from the Pareto front, fall into the inferior zone and thus exhibit low utilities. 
These low-utility solutions are more prominently reflected in the EU metric compared to the Hv and Sp metrics, since they are excluded from computation of Hv and Sp metrics as dominated solutions.
In contrast to PEDA, most solutions of our algorithm for out-of-distribution target preferences remain consistently close to the Pareto front and thus exhibit comparatively high utility.
This outcome demonstrates superior performance stability of our algorithm in dealing with out-of-distribution preferences.

\begin{figure*}[h!]
\begin{center}
    \centerline{\includegraphics[width=\textwidth]{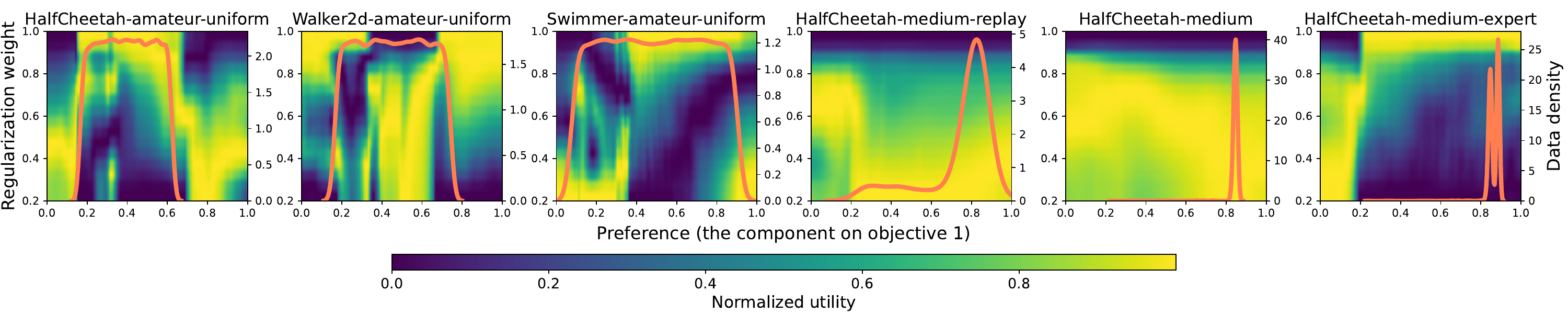}}
    \caption{Relationship between adapted regularization weight and offline data distribution. 
    The orange line represents the data density under different behavior preference in offline dataset.
    The heatmap indicates the utility under various target preferences and different regularization weights.
    }
    \label{fig:wbc_and_data}
\end{center}
\end{figure*}

\subsection{Relationship between adapted regularization weight and offline data distribution}
In Figure~\ref{fig:wbc_and_data}, we present the data density under different behavior preferences using the orange line. 
A higher point on the orange line indicates a larger number of trajectories with the respective behavior preference. 
Additionally, we display a heatmap within the same subplot where brighter areas signify higher utility obtained by using the corresponding regularization weight under the corresponding preference.
We observe a strong correlation between the optimal regularization weight and the distribution of data under various behavior preferences. 
In some environments (e.g., such as HalfCheetah-amateur-uniform, Swimmer-amateur-uniform, and HalfCheetah-medium-expert), a higher data density is associated with a larger optimal regularization weight, while in other environments (e.g., HalfCheetah-medium-replay and Walker2d-mateur-uniform), the opposite trend is observed.

\subsection{Approximate Pareto Front on all Environments}
The approximate Pareto fronts generated from our methods on D4MORL and MOSB datasets are depicted in Figure~\ref{fig:all_d4morl_pareto_front} and Figure~\ref{fig:all_d4rl_pareto_front}, respectively.
The results demonstrate that our method consistently achieves a broadly expanded approximate Pareto front with a dense set of learned policies, using datasets collected from both multi-preference and single-preference behavior policies.

\begin{figure*}[t!]
\vspace{-1cm}
\begin{center}
    \centerline{\includegraphics[width=0.75\textwidth]{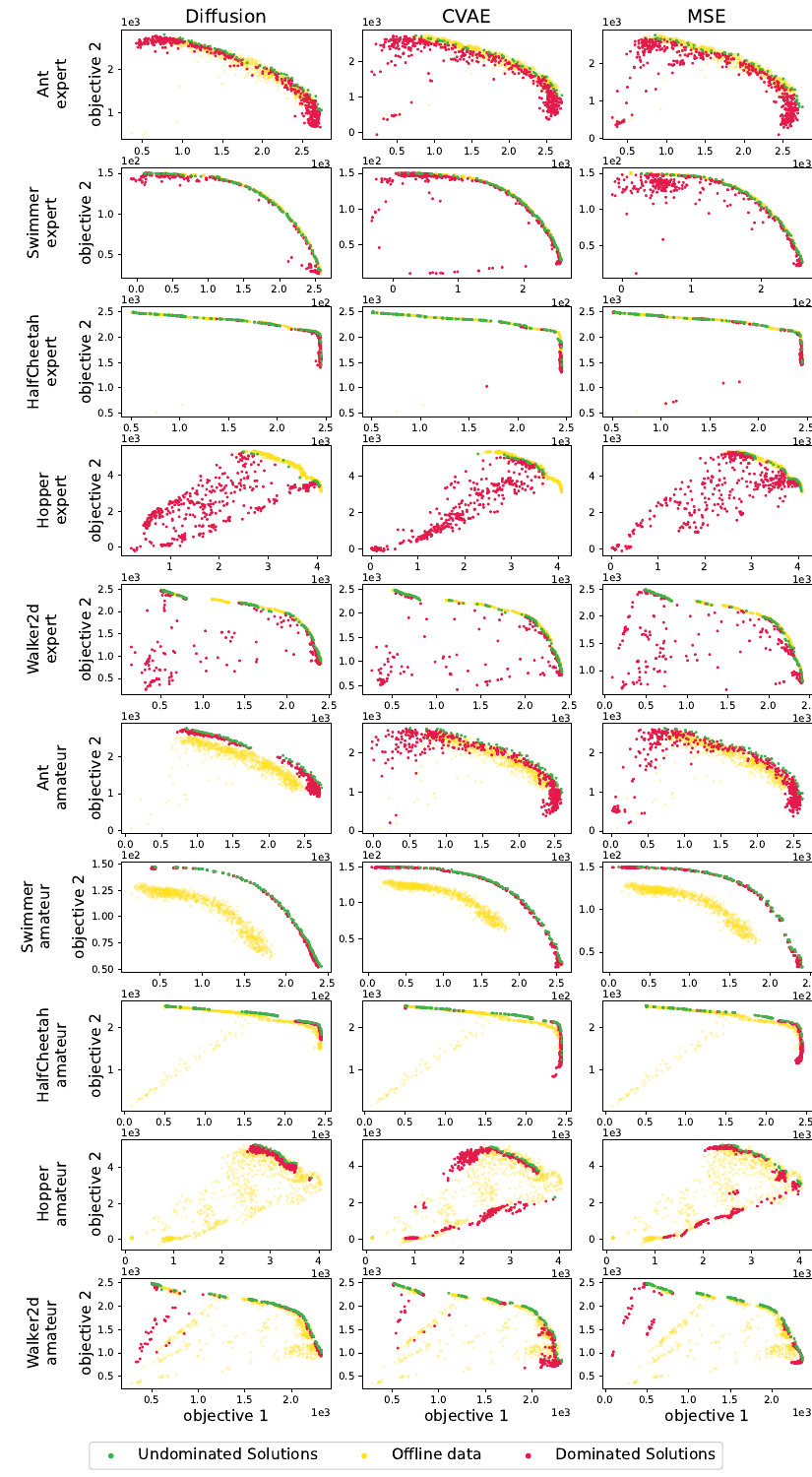}}
    \caption{The approximate Pareto Front obtained by our methods on all D4MORL environments.
    }
    \label{fig:all_d4morl_pareto_front}
\end{center}
\end{figure*}

\begin{figure*}[t!]
\begin{center}
    \centerline{\includegraphics[width=0.75\textwidth]{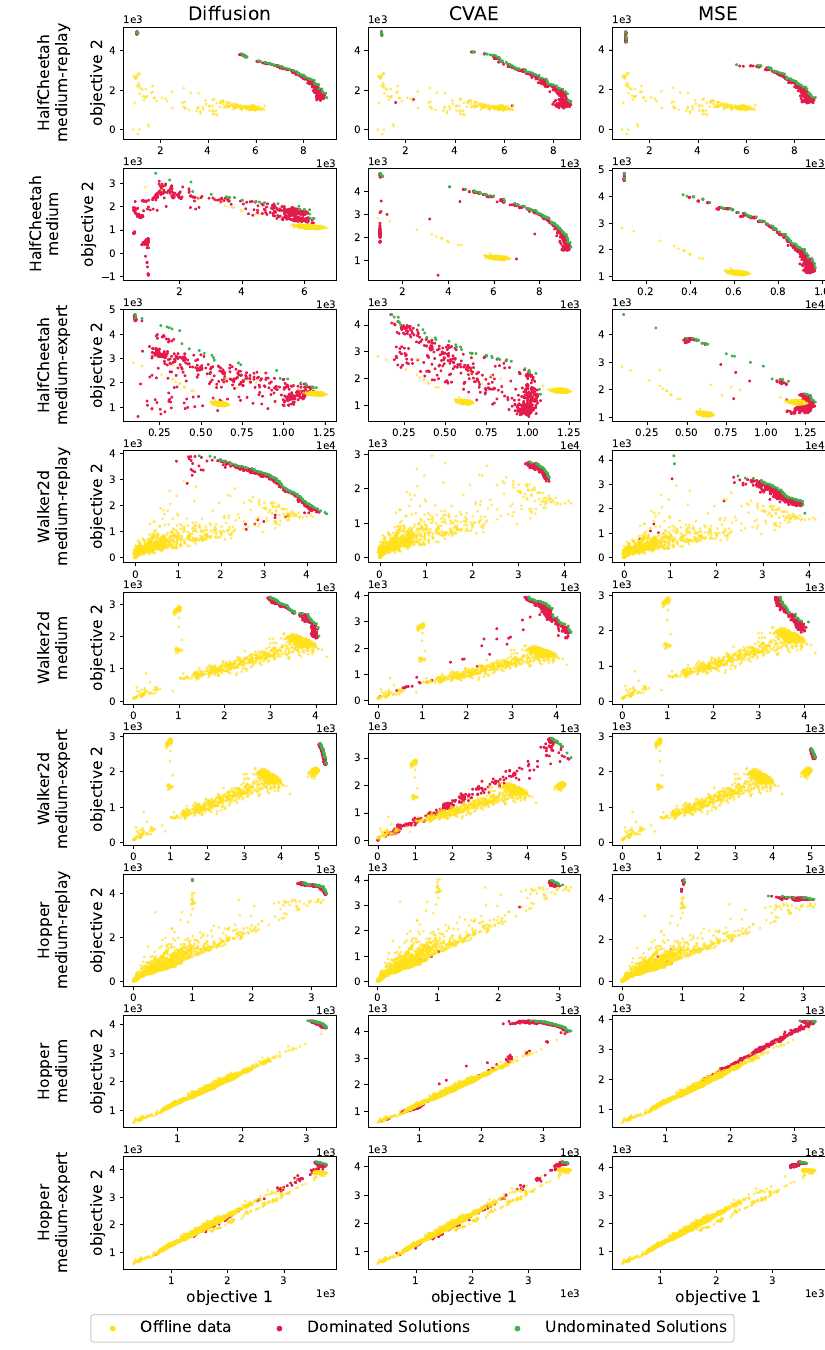}}
    \caption{The approximate Pareto Front obtained by our methods on all MOSB environments.
    }
    \label{fig:all_d4rl_pareto_front}
\end{center}
\end{figure*}

\fi


\end{document}